\documentclass{article}

\PassOptionsToPackage{numbers, compress}{natbib}
\usepackage{algorithm}
\usepackage{algorithmic}
\usepackage{lipsum}
    \usepackage[final]{neurips_2024}

\usepackage[utf8]{inputenc} % allow utf-8 input
\usepackage[T1]{fontenc}    % use 8-bit T1 fonts
\usepackage{hyperref}       % hyperlinks
\usepackage{url}            % simple URL typesetting
\usepackage{booktabs}       % professional-quality tables
\usepackage{amsfonts}       % blackboard math symbols
\usepackage{nicefrac}       % compact symbols for 1/2, etc.
\usepackage{microtype}      % microtypography
\usepackage{xcolor}         % colors

\usepackage{microtype}
\usepackage{graphicx}
\usepackage{wrapfig}
\usepackage{caption}

\usepackage{subfigure}
\usepackage{booktabs} % for professional tables

\usepackage{tabularray} 
\UseTblrLibrary{booktabs}
\usepackage{subcaption}
\usepackage{multirow}

\usepackage{xcolor, soul}
\definecolor{lightblue}{rgb}{0.08,0.08,0.95}

\newcommand{\blue}[1]{{#1}} % for submission

\newcommand{\diff}{\mathop{}\!\mathrm{d}}
\newcommand{\expp}{\mathrm{e}}
\newcommand{\cond}{{\;|\;}}

\usepackage{hyperref}

\usepackage{amsmath}
\usepackage{amssymb}
\usepackage{mathtools}
\usepackage{amsthm}
\usepackage{bm}

\usepackage[utf8]{inputenc} % allow utf-8 input
\usepackage[T1]{fontenc}    % use 8-bit T1 fonts
\usepackage{url}            % simple URL typesetting
\usepackage{amsfonts}       % blackboard math symbols
\usepackage{nicefrac}       % compact symbols for 1/2, etc.
\usepackage[capitalize,noabbrev]{cleveref}

\theoremstyle{plain}
\newtheorem{theorem}{Theorem}[section]
\newtheorem{proposition}[theorem]{Proposition}

\theoremstyle{definition}

\theoremstyle{remark}

\newcommand{\Update}[1]{{\color{black}#1}}

\usepackage[textsize=tiny]{todonotes}

\title{Entropy-regularized Diffusion Policy with Q-Ensembles for Offline Reinforcement Learning}

\author{%
  Ruoqi Zhang\thanks{Corresponding authors} \quad Ziwei Luo\footnotemark[1] \quad Jens Sj{\"o}lund \quad Thomas B. Sch{\"o}n \quad Per Mattsson\\
  Department of Information Technology, Uppsala University\\
  \texttt{\{ruoqi.zhang,ziwei.luo,jens.sjolund,thomas.schon,per.mattsson\}@it.uu.se}
}

\begin{document}

\maketitle

\begin{abstract}

Diffusion policy has shown a strong ability to express complex action distributions in offline reinforcement learning (RL). However, it suffers from overestimating Q-value functions on out-of-distribution (OOD) data points due to the offline dataset limitation. To address it, this paper proposes a novel entropy-regularized diffusion policy and takes into account the confidence of the Q-value prediction with Q-ensembles. At the core of our diffusion policy is a mean-reverting stochastic differential equation (SDE) that transfers the action distribution into a standard Gaussian form and then samples actions conditioned on the environment state with a corresponding reverse-time process. We show that the entropy of such a policy is tractable and that can be used to increase the exploration of OOD samples in offline RL training. Moreover, we propose using the lower confidence bound of Q-ensembles for pessimistic Q-value function estimation. The proposed approach demonstrates state-of-the-art performance across a range of tasks in the D4RL benchmarks, significantly improving upon existing diffusion-based policies. The code is available at \href{https://github.com/ruoqizzz/entropy-offlineRL}{https://github.com/ruoqizzz/entropy-offlineRL}.

\end{abstract}

\section{Introduction}
 Offline reinforcement learning (RL), also known as batch RL \citep{lange2012batchrl} focuses on learning optimal policies from a previously collected dataset without further active interactions with the environment  \citep{levine2020offline}. 
Although offline RL offers a promising avenue for deploying RL in real-world settings where online exploration is infeasible, 
a key challenge lies in deriving effective policies from fixed datasets, which usually are diversified and sub-optimal. The direct application of standard policy improvement approaches is hindered by the distribution shift problem \citep{fujimoto2019off}. Previous works mainly address this issue by either regularizing the learned policy close to the behavior policy \citep{fujimoto2019off, fujimoto2021minimalist} or by making conservative updates for Q-networks \citep{kumar2020conservative, kostrikov2021IQL}.

Diffusion models have rapidly become a prominent class of highly expressive policies in offline RL~\citep{wang2022diffusionpolicy,zhu2023diffusion}.
While this expressiveness is beneficial when modeling complex behaviors, it also means that the model has a higher capacity to overfit the noise or specific idiosyncrasies in the training data. To address this, existing work introduce Q-learning guidance and regard the diffusion loss as a special regularizer adding to the policy improvement process~\citep{wang2022diffusionpolicy,hansen2023idql,kang2023efficient}. Such a framework has achieved impressive results on offline RL tasks. However, its performance is limited by pre-collected datasets (or behavior policies) and the learning suffers severe overestimation of Q-value functions on unseen state-action samples \citep{levine2020offline}.

\begin{figure}[t]
\begin{center}
\centerline{\includegraphics[width=\columnwidth]{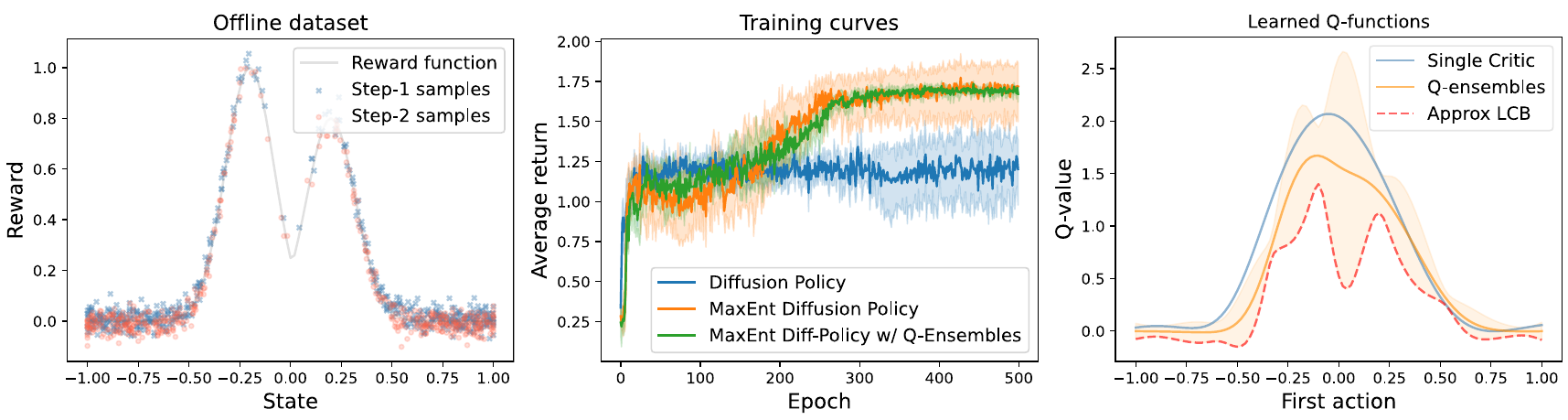}}
% \vspace{-0.5em}
\caption{A toy RL task in which the agent sequentially takes two steps (starting from 0) to seek a state with the highest reward. \textbf{Left}: The reward function is a mixture of Gaussian, and the offline data distribution is unbalanced with most samples located in low-reward states. \textbf{Center}: Training different policies on this task with 5 random seeds for 500 epochs. We find that a diffusion policy with entropy regularization and Q-ensembles yields the best results with low training variance. \textbf{Right}: Learned Q-value curve for the first step actions in state 0. The approximation of the lower confidence bound (LCB) of Q-ensembles is also plotted.}
\label{fig:teaser}
\end{center}
% \vskip -0.4in
% \vspace{-3em}
\end{figure}

One promising approach is to increase exploration for out-of-distribution (OOD) actions, with the hope that the RL agent can be more robust to diverse Q-values and estimation errors~\citep{ziebart2010modeling}. Previous online RL algorithms achieve this by maximizing the entropy of pre-defined tractable policies such as Gaussians~\citep{mnih2016asynchronous,haarnoja2017reinforcement,haarnoja2018soft}. Unfortunately, directly computing the log probability of a diffusion policy is almost impossible since its generative process is a stochastic denoising sequence. Moreover, it is worth noting that entropy is seldom used in offline settings because it may lead to a distributional shift issue which may cause overestimation of Q-values on unseen actions in the offline dataset. 

Another line of work addresses the overestimation problem by enforcing the Q-values to be more pessimistic~\citep{kumar2020conservative,jin2021pessimism}. 
Inspired by this, uncertainty-driven RL algorithms employ an ensemble of Q-networks to provide different Q-value predictions for the same state-action pairs~\citep{an2021edac,bai2022pessimistic}. 
The variation in these predictions serves as a measure of uncertainty. For state-action pairs exhibiting high predictive variance (e.g., OOD data points), these methods preferentially adopt pessimistic Q-value estimations as policy guidance. 

In this paper, we present an entropy-regularized diffusion policy with Q-ensembles for offline RL. At the core of our method is a mean-reverting stochastic differential equation (SDE)~\cite{luo2023image} which allows us to sample actions from standard Gaussian conditioned on the environment state. We show that such an SDE provides a tractable entropy regularization that can be added in training to increase the exploration of OOD data points. In addition, we approximate the lower confidence bound (LCB) of Q-ensembles to alleviate potential distributional shifts, thereby learning a pessimistic policy to handle high uncertainty scenarios from offline datasets.
As illustrated in Figure~\ref{fig:teaser}, both entropy regularization and Q-ensembles can improve RL performance on unbalanced offline datasets. 
The LCB approach further reduces the variance between different trials and provides a better estimation of unseen state-action pairs. 

Our model achieves highly competitive performance across a range of offline D4RL benchmark tasks~\cite{fu2020d4rl} and, in particular, significantly outperforms other diffusion-based approaches in the Antmaze environment. The superior performance demonstrates the effectiveness of the entropy-regularization and Q-ensembles. Overall, the proposed method encourages policy diversity and cautious decision-making, enhancing exploration while grounding the policy in the confidence of its value estimates derived from the offline dataset.

\section{Background}
\label{sec:background}
This section reviews the core concepts of offline RL and then introduces the mean-reverting SDEs and shows how we sample actions from its reverse-time process. Note that there are two types of timesteps for RL and SDE. To clarify that, we use $i\in \{0,\dots,N\}$ to denote the RL trajectories' step and $t\in\{0,\dots, T\}$ to index diffusion discrete times.

\paragraph{Offline RL.}
We consider learning a Markov decision process (MDP) defined as $M=\{ \mathcal{S}, \mathcal{A}, P, R, \gamma, d_0 \}$, where $\mathcal{S}$ and $\mathcal{A}$ are the state and action spaces, respectively. The state transition probability is denoted $P(\mathbf{s}_{i+1} \cond \mathbf{s}_i, \mathbf{a}_i)$ and $R: \mathcal{S}\times \mathcal{A} \rightarrow \mathbb{R}$ represents a reward function, $\gamma \in (0,1]$ is the discount factor, and $d_0$ is the initial state distribution. The goal of RL is to maximize the cumulative discounted reward $\sum_{i=0}^\infty \gamma^i\mathbb{E}_{\mathbf{a}_i \sim \pi(\mathbf{s}_i)}\big[ r(\mathbf{s}_i, \mathbf{a}_i) \big]$ with a learned policy $\pi$. In contrast to online RL which requires continuous interactions with the environment, offline RL directly learns the policy from the static dataset $\mathcal{D} = \{ (\mathbf{s}_i, \mathbf{a}_i, r_i, \mathbf{s}_{i+1})\}_{i=1}^{N_\mathcal{D}}$. In the offline setting, two primary challenges are frequently encountered: over-conservatism and a limited capacity to effectively utilize diversified datasets~\citep{levine2020offline}. To address the issue of limited capacity, diffusion models have recently been employed to learn complex behavior policies from datasets~\citep{wang2022diffusionpolicy}.

\paragraph{Mean-Reverting SDE.}

Assume that we have a random variable 
$\bm{a}^0$ 
sampled from an unknown distribution 
$p_0(\bm{a})$
. 
The mean-reverting SDE 
\citep{luo2023image} is a diffusion process $\{\bm{a}^t\}_{t \in [0,T]}$ that gradually injects noise to $\bm{a}^0$:

\begin{equation}
    \diff \bm{a} = -{\theta_t} \bm{a} \diff t + \sigma_t \diff \bm{w}, \quad \bm{a}^0 \sim p_0(\bm{a}),
    \label{eq:t-sde}
\end{equation}
where $\bm{w}$ is the standard Wiener process, $\theta_t$ and $\sigma_t$ are predefined positive parameters that characterize the speed of mean reversion and the stochastic volatility, respectively. Compared to IR-SDE~\citep{luo2023image}, we set the mean to 0 to let the process drift to pure noise to fit the RL environment. 
The mean can however be tuned to high-reward actions in the offline dataset or prior knowledge.
By setting $\sigma_t^2 = 2\theta_t$ for all diffusion steps, the solution to the forward SDE ($\tau < t$) is given by

\begin{equation}
    p(\bm{a}^t \cond \bm{a}^\tau) = 
    \mathcal{N}(\bm{a}^t \cond \bm{a}^\tau \expp^{-\bar{\theta}_{\tau:t}}, (1 - \expp^{-2 \, \bar{\theta}_{\tau:t}}) \bm{I}),
    \label{eq:sde-solution}
\end{equation}
where $\bar{\theta}_{\tau:t} \coloneqq \int^{t}_\tau \theta_z \diff z$ are known coefficients~\citep{luo2023image}. In the limit $t \to \infty$, the marginal distribution
% $p_t(\bm{x}) = p(\bm{x}_t \cond \bm{x}_0)$ 
$p_t(\bm{a}) = p(\bm{a}^t \cond \bm{a}^0)$ 
converges to a standard Gaussian $\mathcal{N}(0, \bm{I})$. This gives the forward process its informative name, i.e. ``\textit{mean-reverting}''.
Then, \citet{anderson1982reverse} states that we can generate new samples from Gaussian noises by reversing the SDE~\eqref{eq:t-sde} as

\begin{equation}
    \diff {\bm{a}} = \big[ -\theta_t \, \bm{a} - \sigma_t^2 \, \nabla_{\bm a} \log p_t({\bm a}) \big] \diff t + \sigma_t \diff \bar{\bm w}, 
    \label{eq:reverse-sde}
\end{equation}
where $\bm{a}^T \sim \mathcal{N}(0, {\bm I})$ and $\bar{\bm w}$ is the reverse-time Wiener process. This reverse-time SDE provides a strong ability to fit complex distributions, such as the policy distribution represented in the dataset $\mathcal{D}$. 
Moreover, the ground truth score 
$\nabla_{\bm a} \log p_t({\bm a})$ is acquirable in training. We can thus combine it with the reparameterization trick

\begin{equation}
     \bm{a}^t = \bm{a}^0 \expp^{-\bar{\theta}_{t}} + \sqrt{1 - \expp^{-2\bar{\theta}_{t}}} \cdot \bm{\epsilon}_t
    \label{eq:reparameterize_xt}
\end{equation}
to train a time-dependent neural network ${\bm \epsilon}_\phi$ using noise matching on randomly sampled timesteps:
\begin{equation}
    L_\text{diff}(\phi) \coloneqq \mathbb{E}_{t \in [0, T]} \Big[ \big\lVert {\bm{\epsilon}}_\phi(\bm{a}^t, t) - {\bm \epsilon}_t) \bigr\rVert\Big],
    \label{eq:noise_objective}
\end{equation}
where $\bm{\epsilon}_t \sim \mathcal{N}(0, \bm{I})$ is a Gaussian noise and 
$\{{\bm a}^t\}_{t=0}^T$ 
denotes the discretization of the diffusion process.
See Appendix \ref{prf:SDE_solution} for more details about the solution, reverse process, and loss function.

\paragraph{Sample Actions with SDE.}
Most existing RL algorithms employ unimodal Gaussian policies with learned mean and variance. 

However, this approach encounters a challenge when applied to offline datasets, which are typically collected by a mixture of policies and therefore hard to represent by a simple Gaussian model. 
Thus we prefer to represent the policy with an expressive model such as the reverse-time SDE. More specifically, the forward SDE provides theoretical guidance to train the neural network, and the reverse-time SDE \eqref{eq:reverse-sde} generates actions from Gaussian noise conditioned on the current environment state as a typical score-based generative process~\citep{song2020score}.

\section{Method}

We present the three core components of our method: 1) an efficient sampling strategy based on the mean-reverting SDE; 2) an entropy regularization term that enhances action space exploration; and 3) a pessimistic evaluation with Q-ensembles that avoids overestimation of unseen actions. 

\subsection{Optimal Sampling with Mean-Reverting SDE}
We have shown how to sample actions with reverse-time SDEs in Section~\ref{sec:background}. 
However, generating data from the standard mean-reverting SDE~\citep{luo2023image} requires many diffusion steps and is sensitive to the noise scheduler~\citep{nichol2021improved}.
To improve sample efficiency, we propose generating actions from the posterior distribution $p({\bm a}^{t-1} \cond {\bm a}^t)$ conditioned on ${\bm a}^0$. This approach ensures fast convergence of the generative process while preserving its stochasticity.

\begin{proposition}
\label{prop:posterior}
    Given an initial variable ${\bm a}^0$, for any diffusion state ${\bm a}^t$ at time $t \in [1,T]$, the posterior of the mean-reverting SDE \eqref{eq:t-sde} conditioned on ${\bm a}^0$ is 

    \begin{equation}
        p({\bm a}^{t-1} \cond {\bm a}^t, {\bm a}^0) = \mathcal{N}({\bm a}^{t-1} \cond \tilde{\mu}_t({\bm a}^t, \, {\bm a}^0), \; \tilde{\beta}_t \mathbf{I}),
    \label{eq:posterior}
    \end{equation}
    which is a Gaussian with mean and variance given by:

    \begin{equation}
    \begin{split}
        \tilde{\mu}_t({\bm a}^t, {\bm a}^0) \coloneqq \frac{1 - \expp^{-2\bar{\theta}_{t-1}}}{1 - \expp^{-2\bar{\theta}_{t}}}{\expp^{-\theta^{'}_t}} {\bm a}^t + \frac{1 - \expp^{-2\theta^{'}_t}}{1 - \expp^{-2\bar{\theta}_{t}}}{\expp^{-\bar{\theta}_{t-1}}} {\bm a}^0 
        % \\[2mm]
        % \mathrm{and} 
         \quad \text{and} \quad 
        \tilde{\beta}_t \coloneqq \frac{(1 - {\expp^{-2\bar{\theta}_{t-1}}})(1 - {\expp^{-2\theta^{'}_t}})}{1 - \expp^{-2\bar{\theta}_{t}}},
        \label{eq:posterior_mu_var}
    \end{split}
    \end{equation}
    where $\theta_i^{'} \coloneqq \int_{i-1}^i \theta_t dt$ and $\bar{\theta}_{t}$ is to substitute  $\bar{\theta}_{0:t}$ for clear notation.
\end{proposition}
The proof is provided in Appendix \ref{prf:posterior}. Moreover, thanks to the reparameterization trick~\citep{kingma2013auto}, we can approximate the variable ${\bm a}_0$ by reformulating Eq.~\eqref{eq:reparameterize_xt} to

\begin{equation}
    \hat{\bm a}^0
    = \expp^{\bar{\theta}_{t}} 
        \big ({\bm a}^t - \sqrt{1-\expp^{-2\bar{\theta}_{t}}} \bm{{\epsilon}}_\phi({\bm a}^t, \, t)
        \big),
    \label{eq:est_x0}
\end{equation}
where $\bm{{\epsilon}}_\phi$ is the learned noise prediction network. Then we combine Eq.~\eqref{eq:est_x0} with Eq.~\eqref{eq:posterior} to iteratively construct the sampling process. In addition, it can be proved that the distribution mean
\begin{wrapfigure}{r}{0.6\textwidth}
\vspace{-0.5em}
\centering
% \begin{figure}[t]
\centerline{\includegraphics[width=0.55\columnwidth]{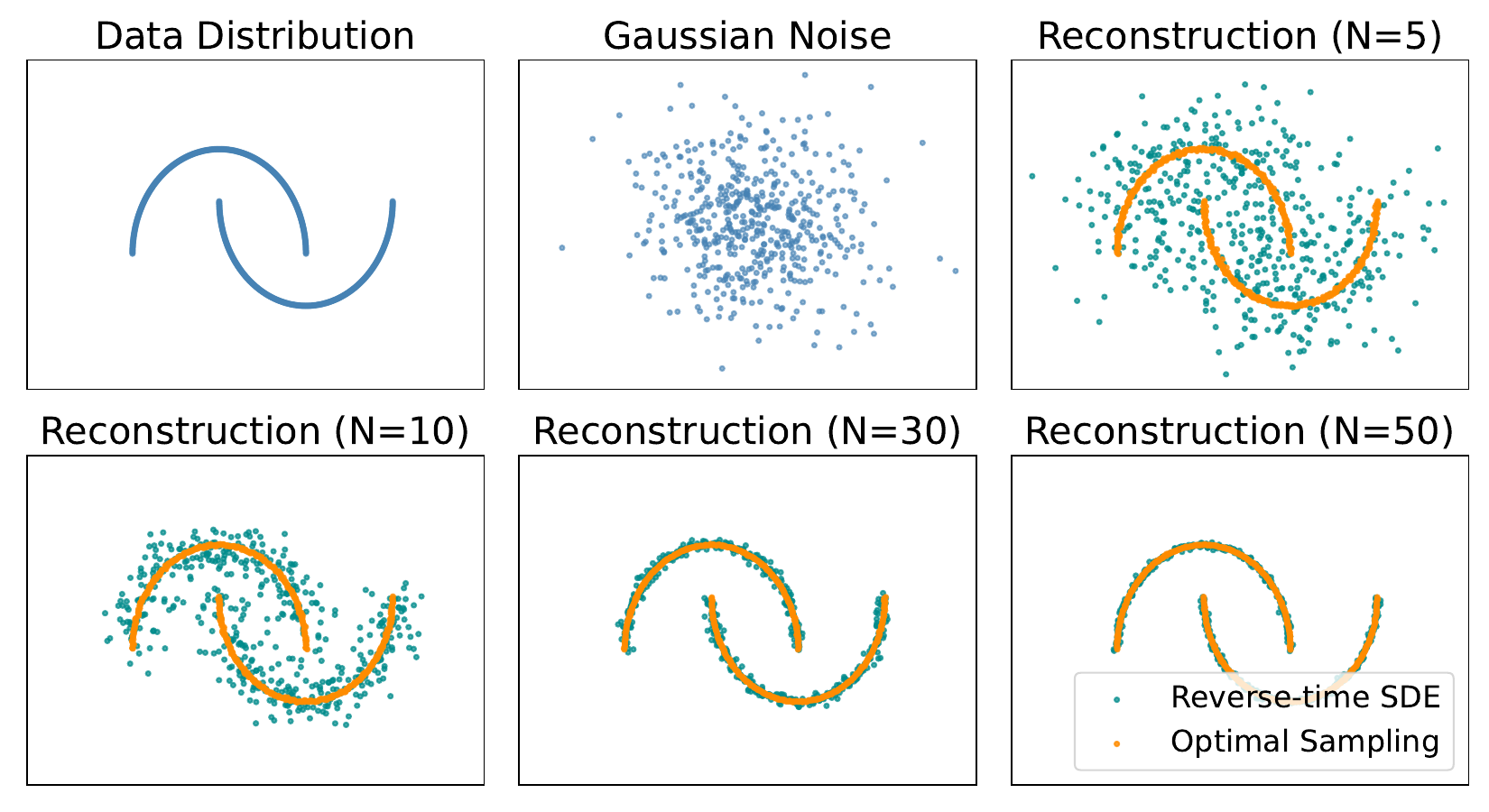}}
\vspace{-2.0mm}
\caption{Comparison of the reverse-time SDE and optimal sampling process in data reconstruction.}
\label{fig:sde_vs_posterior}
\vspace{-1.8em}
\end{wrapfigure}
$\tilde{\mu}_t({\bm a}^t, {\bm a}^0)$ 
is the optimal reverse path from 
${\bm a}^t$ to ${\bm a}^{t-1}$ 
(see Appendix \ref{prf:optimal_reverse_path}). Here, we illustrate a simple example of data reconstruction with different diffusion steps in Figure~\ref{fig:sde_vs_posterior}.
\blue{Additional results are shown are provided in Appendix~\ref{app-sec:optimal_sampling}.
}
The proposed optimal sampling process from Proposition~\ref{prop:posterior} requires only 5 steps, versus over 30 for the standard process. It clearly shows that the proposed optimal sampling is more efficient than the standard reverse-time SDE process. 

% \vspace{0.5em}
\paragraph{Notation Note} Recall that we have two distinct types of timesteps for RL and SDE denoted by $i$ and $t$, respectively. To clarify the notation, in the following sections, we use ${\bm a}^t_i$ to represent the intermediate variable of an action taken at RL trajectory step $i$ with SDE timestep $t$, as ${\bm a}^t_i = {\bm a}^t$ at state ${\bm s}_i$.

Therefore, the action to take for state ${\bm s}_i$ is the final sampled action ${\bm a}_i$ denoted by ${\bm a}^0_i$. 
Hence, the policy is given by 
\begin{equation}
    \pi_\phi({\bm a}^0_i \cond {\bm s}_i) = p_\phi({\bm a}^{0})
\end{equation}
While we cannot sample directly from this distribution we can efficiently sample the SDE's reverse joint distribution as

\footnotesize
\begin{equation}
    p_\phi({\bm a}^{0:T})=p({\bm a}^T)\prod_{i=1}^T p_\phi({\bm a}^{t-1} \cond {\bm a}^{t}),
    \label{eq:joint_distribution}
\end{equation}
\normalsize
where $p({\bm a}^T) = \mathcal{N}(0, {\bm I})$ 
is Gaussian noise and the generative process is conditioned on the environment state ${\bm s}_i$. So to take an action from  $\pi_\phi({\bm a}^0_i \cond {\bm s}_i)$, we sample from the joint distribution using Eq.~\eqref{eq:posterior} and Eq.~\eqref{eq:est_x0} and finally pick out 
${\bm a}_0$
as our sampled action. 
\blue{
The visualization of is out method is provided in Appendix~\ref{app-sec:visualization}.
}

\subsection{Diffusion Policy with Entropy Regularization}

The simplest strategy of learning a diffusion policy is to inject Q-value function guidance to the noise matching loss~\eqref{eq:noise_objective}, in the hope that the reverse-time SDE~\eqref{eq:reverse-sde} would learn to sample actions with higher values. This can be easily achieved by minimizing the following objective:
\begin{equation}
    J_\pi(\phi) = L_\text{diff}(\phi) - \mathbb{E}_{{\bm s}_i \sim \mathcal{D}, {\bm a}^0_i \sim \pi_\phi}\left[Q_{\psi}({\bm s}_i, {\bm a}_i^0) \right],
\label{eq:diffusion-policy}
\end{equation}
where $Q_{\psi}$ is the state-action value function approximated by a neural network, see Section~\ref{sec:qensembles}.

This combination regards diffusion loss as a behavior-cloning term that learns the overall action distribution from offline datasets. However, the training is limited to existing data samples and the Q-learning term is sensitive to unseen actions. To address it, we propose to add an additional entropy term $\mathcal{H} = \mathbb{E}_{{\bm s}_i \sim \mathcal{D}} \left[ -\log \pi_\phi(\cdot \cond {\bm s}_i) \right]$ to increase the exploration of the action space during training and rewrite the policy loss  \eqref{eq:diffusion-policy} to
\begin{equation}
\begin{split}
    J_\pi(\phi) = \, L_\text{diff}(\phi) - \lambda \, \mathbb{E}_{{\bm s}_i \sim \mathcal{D}, {\bm a}^0_i \sim \pi_\phi}\left[Q_{\psi}({\bm s}_i, {\bm a}_i^0) - \alpha \log \pi_\phi({\bm a}_i^0 \cond {\bm s}_i) \right].
    \label{eq:maxen-diffusion-policy}
\end{split}
\end{equation}
where $\alpha$ is a hyperparameter that determines the relative importance of the entropy term versus Q-values, and $\lambda=\eta \,/\, {\mathbb{E}_{(s,a)\sim \mathcal{D}} [\vert Q_\psi(s,a)\vert]}$ to normalize the scale of the Q-values and balance loss terms. Iteratively generating the action ${\bm a}_i^0$ though a reverse diffusion process is computationally costly but, with an estimated noise $\bm{{\epsilon}}_\phi$ from diffusion term~\eqref{eq:noise_objective}, we can thus directly use it to approximate ${\bm a}_i^0$ based on Eq.~\eqref{eq:est_x0} for more efficient training. 

\paragraph{Entropy Approximation.} It is worth noting that the log probability of the policy $\log (\pi_\phi ({\bm a}_i^0 \cond {\bm s}_i))$ is in general intractable in the diffusion process. However, we found that the log probability of the joint distribution in Eq.~\eqref{eq:joint_distribution} is tractable when conditioned on the sampled action ${\bm a}_i^0$.  
Proposition~\ref{prop:posterior} further shows that the conditional posterior from ${\bm a}_i^1$ to ${\bm a}_i^0$ is Gaussian, meaning that 
\begin{equation}
    -\log\pi_\phi({\bm a}_i^0 \cond {\bm s}_i) = -\log\pi_\phi({\bm a}_i^1 \cond {\bm s}_i) + \mathcal{C},
    \label{eq:a0_propto_a1}
\end{equation}
where $\mathcal{C}$ is a constant and ${\bm a}_i^1$ can be approximated using Eq.~\eqref{eq:sde-solution} similar to ${\bm a}_i^0$. The proof is provided in Appendix~\ref{prf:entropy_approximation}.
Then we can focus on the conditional reverse marginal distribution $p_\phi({\bm a}_i^1 \cond {\bm a}_i^T, {\bm s}_i)$ that determines the exploration of actions and is acquirable via Bayes' rule:
\begin{equation}
    p_\phi({\bm a}_i^1 \cond {\bm a}_i^T, s_i) = \frac{ p_\phi({\bm a}_i^T \cond {\bm a}_i^1, {\bm s}_i) \; p_\phi({\bm a}_i^1 \cond {\bm a}_i^0, {\bm s}_i)}{p_\phi({\bm a}_i^T \cond {\bm a}_i^0, {\bm s}_i)}.
    \label{eq:approx_pi_bayes}
\end{equation}
Since all terms in Eq.~\eqref{eq:approx_pi_bayes} can be computed with Eq.~\eqref{eq:sde-solution}, we can rewrite the policy objective as
\small
\begin{equation}
\begin{split}
    J_\pi(\phi) 
    = L_\text{diff}(\phi) 
    - \lambda \, \mathbb{E}_{{\bm s}_i \sim \mathcal{D}, (\hat{\bm a}_i^0, \hat{\bm a}_i^1) \sim \pi_\phi}\big[ Q_{\psi}({\bm s}_i, \hat{\bm a}_i^0) - \alpha \log (p(\hat{\bm a}_i^1 \cond {\bm a}_i^T, {\bm s}_i))) \big],
    \label{eq:diffusion-policy-with-entropy}
\end{split}
\end{equation}
\normalsize
where $\hat{\bm a}_i^0$ and $\hat{\bm a}_i^1$ are approximate values calculated based on samples from the diffusion term. Note that the temperature $\alpha$ usually plays an important role in the maximum entropy RL framework and we thus provide a detailed analysis in Section~\ref{sec:analysis}.

\subsection{Pessimistic Evaluation via Q-ensembles}\label{sec:qensembles}
  
Entropy regularization encourages diffusion policies to explore the action space, reducing the risk of overfitting pre-collected data. However, in offline RL, since the agent cannot collect new data during training, this exploration can lead to inaccuracies in value estimation for unseen state-action pairs~\citep{bai2022pessimistic,ghasemipour2022-msg}. Instead of staying close to the behavior policy and being overly conservative, considering the uncertainty in the value function is an alternative approach.

In this work, we consider a pessimistic variant of a value-based method to manage the uncertainty and risks, i.e., the lower confidence bounds (LCB) with Q-ensembles. 
More specifically, we use an ensemble of Q-functions with independent targets to obtain an accurate LCB of Q-values. Each Q-function is updated based on its own Bellman target without sharing targets among ensemble members \citep{ghasemipour2022-msg}, as follows:
\begin{align}
\begin{aligned}
J_Q(\psi^i) &= \mathbb{E}_{{\bm s}_i, {\bm a}_i, r_i,{\bm s}_{i+1} \sim \mathcal{D}}\left [
    Q_{\psi^m}({\bm s}_i,{\bm a}_i) - y^m(r_i, {\bm s}_{i+1}, \pi_\phi)
    \right
    ] \\
    y^m &= r_i + \gamma \mathbb{E}_{{\bm a}_{i+1}\sim \pi_\phi}
    [Q_{\bar{\psi}^m}({\bm s}_{i+1}, {\bm a}_{i+1})]
\end{aligned}
\label{eq:q-update}
\end{align}
where $\psi^m, \bar{\psi}^{m}$ are the parameters of the Q network and Q-target network for the $m$th Q-function. 
Then, the pessimistic LCB values are derived by subtracting the standard deviation from the mean of the Q-value ensemble, 
\vspace{-1em}
\begin{equation}
Q^{\text{LCB}}_\psi  =    \mathbb{E}_{\text{ens}}
% _{m\in \{1,\dots, M\}}
\left[ 
    Q_{\psi^m}({\bm s}, {\bm a}) \right] - \beta\left[ \sqrt{\mathbb{V}_{\text{ens}}[Q_{\psi^m}({\bm s}, {\bm a})]}
    \right]
    \label{eq:q-lcb}
\end{equation}
where $\beta\geq 0$ is a hyperparameter determining the amount of pessimism, 
$\mathbb{V}[Q_{\psi^m}]$ is the variance of the ensembles,  
and $m \in \{1, \dots, M\}$ where $M$ the number of ensembles. 
Then,   
$Q^{\text{LCB}}_\psi $ is used in the policy improvement step to balance entropy regularization and ensure robust performance. 
Finally, we use $Q^{\text{LCB}}_\psi $  as the $Q_{\psi}$ to \eqref{eq:diffusion-policy-with-entropy}. 
We summarize our method in Algorithm~\ref{alg:training}.

\section{Experiment}
\label{sec:exp}

In this section, we evaluate our methods on standard D4RL offline benchmark tasks \citep{fu2020d4rl} and provide a detailed analysis of entropy regularization, Q-ensembles, and training stability.

\begin{wrapfigure}{R}{0.54\textwidth}
    \vspace{-2.2em}
    \begin{minipage}{0.54\textwidth}
      \begin{algorithm}[H]
    \caption{Diffusion Policy with Q-Ensembles}
    \label{alg:training}
        \begin{algorithmic}
           \STATE Initialize parameters for $\pi_\phi$, ${\pi}_{\bar{\phi}}$, $\{Q_{{\psi}^m}, Q_{\bar{\psi}^m} \}^M_{m=1}$.
           \FOR{each iteration}
           \STATE Sample mini-batch $\{ (\bm{s}_i, \bm{a}_i, r_i, \bm{s}_{i+1})\}$ from $\mathcal{D}$.
           
           \STATE {\bfseries \# Ensemble-Q learning}
           \STATE Generate ${\bm a}^0_{i+1} \sim \pi_{\bar{\phi}}({\bm a}_{i+1} \cond {\bm s}_i)$ with \eqref{eq:posterior} and \eqref{eq:est_x0}.
           \STATE Update Q-networks $\{Q_{\psi^m} \}^M_{m=1}$ by \eqref{eq:q-update}.
           
           \STATE {\bfseries \# Diffusion policy learning}
           \STATE Sample $\{{\bm a}^t_i\}_{t\in[0,T]}$ from ${\bm a}_i$ with \eqref{eq:reparameterize_xt}.
           \STATE Predict noise and approximate ${\bm a}^0_i, {\bm a}^1_i$ with \eqref{eq:est_x0}.
           \STATE Update policy $\pi_\phi$ by \eqref{eq:diffusion-policy-with-entropy} using $Q_{\psi}^{\text{LCB}}$ from \eqref{eq:q-lcb}.
           \STATE {\bfseries \# Update target networks}
           \STATE $\bar{\phi} \leftarrow \eta \phi + (1-\eta)\bar{\phi}$,
           \STATE $\bar{\psi}^m \leftarrow \eta \psi^m + (1-\eta)\bar{\psi}^m, \, m \in \{1, \dots, M\}$
           \ENDFOR
        \end{algorithmic}
    \end{algorithm}
    \end{minipage}
    \vspace{-1.em}
  \end{wrapfigure}

\subsection{Setup}

\paragraph{Datasets} We evaluate our approach on four D4RL benchmark domains: Gym, AntMaze, Adroit, and Kitchen. 
In Gym, we examine three robots (halfcheetah, hopper, walker2d) across sub-optimal (medium), near-optimal (medium-expert), and diverse (medium-replay) datasets. 
The AntMaze domain challenges a quadrupedal ant robot to navigate mazes of varying complexities. The Adroit domain focuses on high-dimensional robotic hand manipulation, using datasets from human demonstrations and robot-imitated human actions. Lastly, the Kitchen domain explores different tasks within a simulated kitchen. These domains collectively provide a comprehensive framework for assessing RL algorithms across diverse scenarios.

\paragraph{Implementation Details} 
Following Diffusion-QL \citep{wang2022diffusionpolicy}, 
we keep the network structure the same for all tasks with three MLP layers (hidden size 256, Mish activation \citep{misra2019mish}), 
and train models for $2000$ epochs for Gym and $1000$ epochs for others. Each epoch consists of 1000 training steps with a batch size of 256.
We use Adam~\citep{kingma2014adam} to optimize both SDE and the Q-ensembles. Each model is evaluated by 10 trajectories for Gym tasks and 100 trajectories for others. In addition, our model is trained on an A100 GPU with 40GB memory for about 8 hours per task, and results are averaged over five random seeds.

\paragraph{Hyperparameters}
\label{sec:exp:hyper}
We keep key hyperparameters consistent: Q-ensemble size 64, LCB coefficient $\beta=4.0$. \Update{The entropy temperature $\alpha = 0.01$ for Gym and AntMaze tasks and automated for Adroit and Kitchen tasks.} 
\blue{The SDE sampling step is set to $T=5$ for Gym and Antmaze tasks, $T=10$ for Adroid and Kitchen tasks. }
For 'medium' and 'large' datasets of AntMaze, we use max Q-backup following \citet{wang2022diffusionpolicy} and \citet{kumar2020conservative}. 
We also introduce the maximum likelihood loss for SDE training as proposed by \citet{luo2023image}. More details are in Appendix~\ref{appendix:hyper}.

\subsection{Comparison with other Methods}

We compare our method with extensive baselines for each domain to provide a thorough evaluation
and to understand the contributions of different components in our approach.
The most fundamental among these are the behavior cloning (BC) method, BCQ \citep{fujimoto2019off} and BEAR \citep{kumar2019stabilizing} which restrict the policy to dataset behavior, highlighting the need for policy regularization and exploration.
We also assess against Diffusion-QL \citep{wang2022diffusionpolicy} which integrates a diffusion model for policy regularization guided by Q-values.
This comparison isolates the benefits of our enhanced sampling process and Q-ensemble integration.
Our comparison includes CQL \citep{kumar2020conservative} and IQL \citep{kostrikov2021IQL}, known for conservative Q-value updates. 
Additionally, we consider EDP \citep{kang2023edp}, a variant of IQL with an efficient diffusion policy, and IDQL \citep{hansen2023idql}, which combines IQL as a critic with behavior cloning diffusion policy reweighted by learned Q-values. These comparisons evaluate the effectiveness of integrating diffusion policies with conservative value estimation.
Finally, we include MSG \citep{ghasemipour2022-msg}, which combines independent Q-ensembles with CQL, and DT \citep{chen2021DT}, treating offline RL as a sequence-to-sequence translation problem. These baselines help assess the robustness and generalizability of our method across different approaches.
The performance comparison between baselines and ours is reported in Table~\ref{table:overall_res_diffT} (Gym, Adroit and Kitchen) and Table~\ref{table:antmaze_res} (AntMaze). Detailed results are discussed below.

\paragraph{Gym tasks}
Most approaches perform well on Gym `medium-expert' and `medium-replay' tasks with high-quality data but drop severely on `medium' tasks with suboptimal trajectories.
Diffusion-QL~\citep{wang2022diffusionpolicy} achieves a better performance through a highly expressive diffusion policy. Our method further improves performance across all three `medium' tasks. 
The results illustrate the efficacy of combining diffusion policy with entropy regularization and Q-ensembles in preventing overfitting to suboptimal behaviors. 
By maintaining policy stochasticity, our algorithm encourages the exploration of action spaces, potentially discovering better strategies than those in the dataset.

\paragraph{Adroit and Kitchen}

Most offline approaches cannot achieve expert performance on these tasks due to the narrowness of human demonstrations in Adroit and the indirect, multitask data in Kitchen~\citep{wang2022diffusionpolicy}. 
\blue{Our method outperforms all other approaches in the Kitchen tasks which suggests its ability to “stitching” the dataset and generalization.
In addition, we fix the entropy coefficient $\alpha$ to be the same as other tasks for a robust setting. Even so, our method still achieves a competitive performanc in Adroit tasks.  
This fixed $\alpha$ leads the agent to continuously explore the action space throughout the entire training process, even when encountering unseen states.
While exploration is generally advantageous, it can be detrimental in environments with limited data variability. Additionally, unlike in antmaze tasks, random actions are more likely to negatively impact performance in tasks  where precise control is essential like Adroit.
}
Moreover, it's worth noting that slightly tuning $\alpha$ leads to a SOTA performance, as illustrated in~\Cref{table:alpha-am}.

\begin{table}[t]
    \centering

% \begin{table}[t]
\begin{minipage}{1.\linewidth}
% \vspace{-1em}
\centering
\scriptsize
\caption{Average normalized scores on D4RL benchmark tasks. Results of BC, CQL, IQL, and IQL+EDP are taken directly from \citet{kang2023edp}, and all other results are taken from their original papers. Our results are reported by averaging 5 random seeds.}
\label{table:overall_res_diffT}
\centering
\vspace{0.5em}
% \begin{center}
% \begin{small}
% \begin{sc}
\resizebox{1.\linewidth}{!}{
\begin{tabular}{l|ccccccc|c}
\toprule
Gym Tasks & BC  & DT  & CQL  & IQL  & IDQL-A  & IQL+EDP   & Diff-QL&   Ours \\
\midrule
Halfcheetah-medium-v2 & 42.6 & 42.6 & 44.0   & 47.4  & 51.0    & 48.1 & 51.1 &   \textbf{54.9} \\
Hopper-medium-v2  & 52.9 & 67.6 & 58.5 & 66.3 & 65.4    & 63.1 & 90.5    &\textbf{94.2} \\
Walker2d-medium-v2    & 75.3 & 74.0 & 72.5 & 78.3 & 82.5    & 85.4 & 87.0  &    \textbf{92.5}\\
Halfcheetah-medium-replay-v2& 36.6 & 36.6 & 45.5 & 44.2 & 45.9    & 43.8 & 47.8   &\textbf{57.0} \\
Hopper-medium-replay-v2 & 18.1 & 82.7 & 95.0   & 94.7 & 92.1    & 99.1 & 101.3 &  \textbf{102.7} \\
Walker2d-medium-replay-v2   & 26.0 & 66.6   & 77.2 & 73.9 & 85.1    & 84.0   & \textbf{95.5}   & 94.2 \\
Halfcheetah-medium-expert-v2& 55.2 & 86.8 & 91.6 & 86.7 & 95.9    & 86.7 & \textbf{96.8}   & 90.3 \\
Hopper-medium-expert-v2 & 52.5 & 107.6 & 105.4  & 91.5 & 108.6    & 99.6 & 111.1 &  \textbf{111.9} \\
Walker2d-medium-expert-v2   & 107.5 & 108.1  & 108.8  & 109.6  & \textbf{112.7}   & 109.0 & 110.1  & 111.2 \\
\midrule
\textbf{Average} & 51.9 & 74.7 & 77.6& 77.0&   82.1   & 79.9& 88.0  & \textbf{89.9} \\
\bottomrule
\toprule
Adroit Tasks & BC  & BCQ  & BEAR & CQL  & IQL   & IQL+EDP   & Diff-QL&   Ours\\
\midrule
% Adroit  \\
Pen-human-v1   & 63.9 & 68.9 & -1.0 & 37.5 & 71.5  & 72.7 & \textbf{72.8 }  & 70.0  \\
Pen-cloned-v1  & 37.0 & 44.0 & 26.5 & 39.2 & 37.3  & \textbf{70.0 }  & 57.3   & 68.4 \\
\midrule
\textbf{Average} & 50.5 & 56.5 & 12.8 & 38.4 & 54.4  & \textbf{71.4}  & 65.1  & 69.2\\
\bottomrule
\toprule
Kitchen Tasks & BC  & BCQ  & BEAR & CQL  & IQL   & IQL+EDP   & Diff-QL&   Ours\\
\midrule
% Kitchen \\
kitchen-complete-v0   & 65.0 & 8.1 & 0.0 & 43.8 & 62.5  & 75.5 & 84   & \textbf{92.7} \\
kitchen-partial-v0    & 38.0 & 18.9 & 13.1 & 49.8 & 46.3  & 46.3 & 60.5  & \textbf{66.3} \\
kitchen-mixed-v0  & 51.5 & 8.1 & 47.2 & 51   & 51  & 56.5 & 62.6 &  \textbf{68.0} \\
\midrule
\textbf{Average} & 51.5 & 11.7 & 20.1 & 48.2 & 53.3  & 59.4& 69.0 &  \textbf{75.7} \\
\bottomrule

\end{tabular}
}
% \end{sc}
% \end{small}
\end{minipage}

\begin{minipage}{1.\linewidth}
% \begin{table}[t]
\centering
\scriptsize
\caption{Average normalized scores on D4RL \textbf{AntMaze} tasks. Results of BC, DT, CQL, IQL, and IQL+EDP are taken directly from \citet{kang2023edp}, and all other results are taken from their original papers. Our results are reported by averaging 5 random seeds.}
\label{table:antmaze_res}
\vspace{0.5em}
% \begin{center}
% \begin{small}
% \begin{sc}
\resizebox{1.\linewidth}{!}{
\begin{tabular}{l|cccccccc|c}
\toprule
AntMaze Tasks & BC  & DT  & CQL  & IQL  & MSG & IDQL-A & IQL+EDP   & Diff-QL&  Ours\\
\midrule
Antmaze-umaze-v0  & 54.6 & 59.2 & 74   & 87.5 & 97.8 & 94.0 & 87.5 & 93.4  &  \textbf{100}\\
Antmaze-umaze-diverse-v0    & 45.6 & 53.0 & 84.0   & 62.2 &\textbf{ 81.8} & 80.2 & 62.2 & 66.2  & 79.8 \\
Antmaze-medium-play-v0  & 0.0  & 0.0  & 61.2 & 71.2 & 89.6 & 84.5 & 71.2 & 76.6  &  \textbf{91.4}\\
Antmaze-medium-diverse-v0   & 0.0 & 0.0    & 53.7 & 70.0   & 88.6 & 84.8 & 70.0   & 78.6  & \textbf{91.6} \\
Antmaze-large-play-v0 & 0.0  & 0.0  & 15.8 & 39.6 & 72.6 & 63.5 & 39.6 & 46.4  & \textbf{81.2} \\
Antmaze-large-diverse-v0    & 0.0 & 0.0    & 14.9 & 47.5 & 71.4 & 67.9 & 47.6 & 56.6  &  \textbf{76.4}\\
\midrule
\textbf{Average} & 16.7 & 18.7 & 50.6 & 63.0   & 83.6 & 79.1 & 63.0& 69.6 &  \textbf{86.7}\\
\bottomrule

\end{tabular}
}
% \end{sc}
% \end{small}
% \end{center}
\vspace{-1em}
\end{minipage}
\end{table}
\normalsize

\paragraph{AntMaze}

AntMaze tasks are more challenging, requiring point-to-point navigation with sparse rewards from sub-optimal trajectories \citep{fu2020d4rl}. As shown in Table~\ref{table:antmaze_res}, traditional behavior cloning methods (BC and DT) get 0 rewards on AntMaze medium and large environments.
Our method shows excellent performance on all the tasks in AntMaze even with large complex maze settings and outperforms other methods by a margin.
The result is not surprising because the entropy regularization incentivizes the policy to explore various sub-optimal trajectories within the dataset and stitch them to find a path toward the goal. 
In tasks with sparse rewards, this can be crucial because it prevents premature convergence to suboptimal deterministic policies.
Additionally, employing the LCB of Q-ensembles 
effectively reduces the risk of  taking low-value actions, 
enabling the development of robust policies. 

In general, employing consistent hyperparameters for each domain, along with fixed entropy temperature $\alpha$, LCB coefficient  $\beta$, and ensemble size $M$ across all tasks, our method not only achieves substantial overall performance but also outperforms prior works in the challenging AntMaze tasks. By the comparison with MSG\citep{ghasemipour2022-msg} (Q-ensemble alone) and Diffusion-QL (Diffusion alone), our method further improves results demonstrating its effectiveness in handling complex environments with sparse rewards by effectively combining suboptimal trajectories to find better solutions via action space exploration.

\subsection{Analysis and Discussion}
\label{sec:analysis}

We first study the core components of our method: entropy regularization and Q-ensemble. 
Then we show that adding both significantly improves the training robustness of diffusion-based policies.

\paragraph{Entropy Regularization}

The core idea of applying entropy regularization in offline RL is to increase the exploration of new actions such that the estimation of Q-functions is more accurate, especially for datasets with unbalanced action distribution such as the toy example in Figure~\ref{fig:teaser}. Here we report the results of training the diffusion policy with different entropy temperatures in Table~\ref{table:alpha-am}. 
It is observed that our method with positive entropy coefficients performs better than that without the entropy term. 
In addition, we can extend our model with an automatic entropy adjustment similar to the work in \cite{haarnoja2018soft2}. This approach is marked as ``\textit{auto}'' in Table~\ref{table:alpha-am}. The results show that auto-tuning the entropy temperature further improves the performance in the Adroit and Kitchen domains. Please refer to Appendix~\ref{app-sec:auto-alpha} for more details.
\begin{wraptable}{r}{0.52\textwidth}
\vspace{-1.em}
\centering
% \scriptsize
\caption{Ablation study on entropy temperatures.}
\label{table:alpha-am}
% \small
\resizebox{1.\linewidth}{!}{
\begin{tabular}{l|cccccc}
\toprule
 Entropy temperature $\alpha$ & 0  & 0.01 & 0.05 & 0.1 & auto \\
\midrule
% Antmaze-umaze-diverse-v0& 78.3 & 79.8& \textbf{84.0} & 79.3 & 81.8\\
Antmaze-medium-play-v0 & 85.7 & \textbf{91.4}& 91 & 88.3 & 92.0\\
Antmaze-medium-diverse-v0& 89.0 & 91.6 & 90.7 &\textbf{93.5} & 90.8 \\
Antmaze-large-play-v0  & 77.7 & 81.2 & 78.3& \textbf{82} & 82.0\\
Antmaze-large-diverse-v0 & 73.7 & 76.4 & 71.3 & \textbf{78.3} & 76.0 \\
% \bottomrule
\midrule
HalfCheetah-Medium-v2  & 53.7 &54.9  &54.0   &\textbf{55.3} & 54.2 \\
Hopper-Medium-v2     & 94.8 &94.2   &93.3   &\textbf{97.7 } &  94.0 \\
Walker2D-Medium-v2    & 89.6   &\textbf{92.5}  &91.6   &89.9& 91.9\\
\midrule
Pen-human-v1    & 60.9  & 67.2 &63.6& 69.8& \textbf{78.5} \\
Pen-cloned-v1      & 57.9  & 66.3 & 61.8& 56.5&  \textbf{79.8}\\
\midrule
Kitchen-complete-v0      &80.6   &82.3 & 77.6&54.4 & \textbf{84.4} \\
Kitchen-Mixed-v0      & 57.0   & 60.2 & 50.8 & 56.5 & \textbf{60.4} \\
\bottomrule
\end{tabular}
}
\vspace{-1.em}
\end{wraptable}

\begin{wraptable}{r}{0.5\textwidth}
\vspace{-1.3em}
\centering
% \scriptsize
\caption{Ablation experiments of our entropy-based diffusion policy with different ensemble sizes on selected AntMaze tasks.}
\label{table:q-ensemble}
% \begin{center}
% \begin{small}
% \begin{sc}
% \vspace{-2mm}
\resizebox{1.\linewidth}{!}{
\begin{tabular}{lccc}
\toprule
   Ensemble Size      & $2$   & $4$    &  $64$ \\
\midrule
Antmaze-medium-play-v0      & $84.0$  & $87.2$     &$\textbf{91.4}$\\
Antmaze-medium-diverse-v0   & $71.8$   & $87.2$     &$\textbf{91.6}$\\
Antmaze-large-play-v0       & $54.2$  & $52.4$ &$\textbf{81.2}$\\ 
Antmaze-large-diverse-v0      & $43.2$  & $69.0$ &
$\textbf{76.4}$\\
\midrule
\textbf{Average}            & 63.3  & 74.0     &\textbf{85.2}\\
\bottomrule
\end{tabular}
}
% \end{sc}
% \end{small}
% \end{center}
\vspace{-1em}
\end{wraptable}

\paragraph{Q-Ensembles}
We evaluate our method under different numbers of Q networks $M\in \{2, 4, 64\}$ in the AntMaze environment to explore the effectiveness of Q-ensembles. 
The results with average performance within 5 different seeds are provided in Table~\ref{table:q-ensemble}. The key observations are 1) As the $M$ increases, the model gets better performance and the training process becomes more stable; 2) The standard deviation in the results decreases as $M$ increases, suggesting larger ensembles not only perform better on average but also provide more reliable and consistent results. 3) While increasing $M$ from 2 to 4 shows a substantial improvement, the performance gains decrease with an even larger size. It is worth noting that other offline RL approaches like Diffusion-QL~\cite{wang2022diffusionpolicy} also adopt two Q networks for training robustness. See Appendix \ref{appedix:experiments} for more detailed results.

\paragraph{LCB coefficients $\beta$}

\begin{figure}
    \centering
        % \vspace{0.1em}
        \includegraphics[width=.8\linewidth]{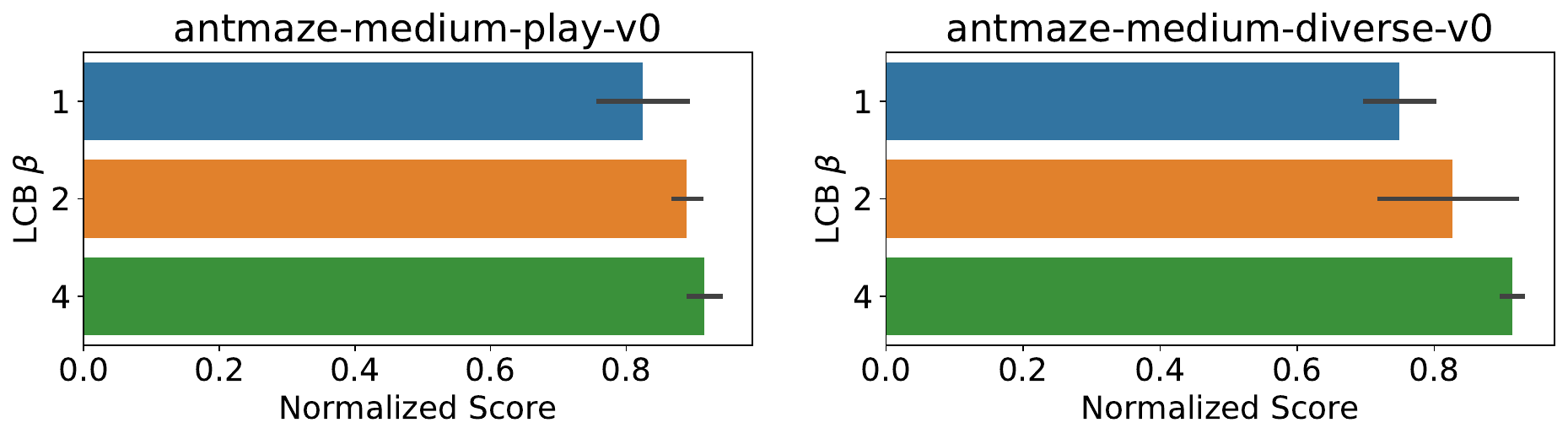}
        \caption{Ablation experiments of our method with different values of LCB coefficient $\beta=1,2,4$ on AntMaze-Medium environments over 5 different random seeds. }
    \label{fig:lcb-beta}
\end{figure}

We evaluate our method with $\beta$ values of 1, 2, and 4 on AntMaze-medium environments 
Figure~\ref{fig:lcb-beta} demonstrates that adjusting the LCB coefficient improves performance, particularly for higher values, which helps in managing the exploration-exploitation trade-off effectively. In addition, the numerical results are provided in Appendix Table~\ref{table:lcb}.

\begin{wrapfigure}{r}{0.55\textwidth}
    \vspace{-0.em}
    \centering
    \footnotesize
    \includegraphics[width=1.\linewidth]{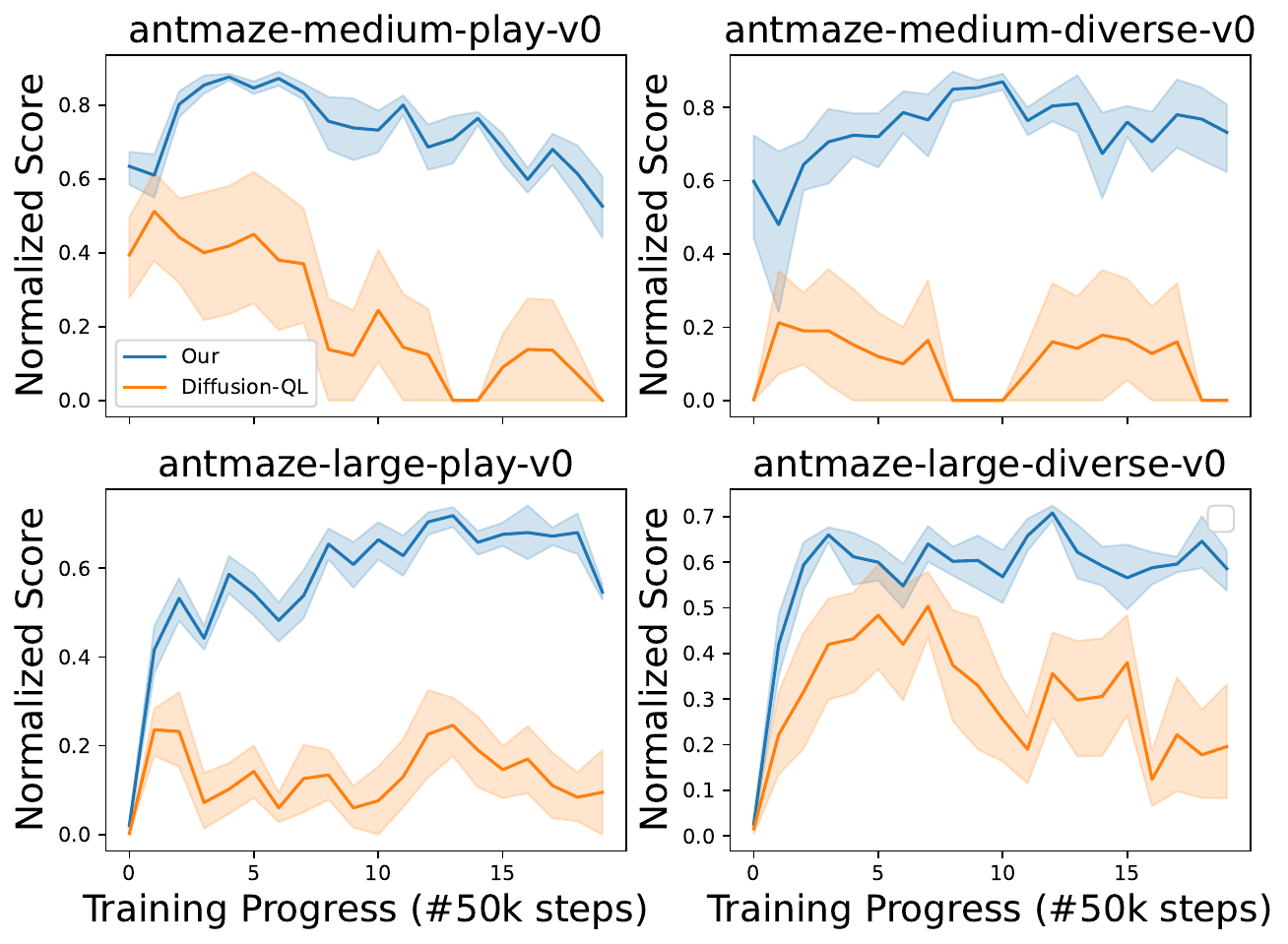}
    \caption{Learning curves of the Diffusion-QL and our method on selected Antmaze tasks over 5 random seeds. }
    \label{fig:ql-eql}
    \vspace{-2em}
\end{wrapfigure}

\paragraph{Training Stability and Computational Time}
Empirically we observe that the training of diffusion policies is always unstable, particularly for sparse-reward environments such as AntMaze medium and large tasks. 
Our method alleviates this problem by incorporating the entropy regularization and Q-ensembles as stated in the introduction. 
Here, we further show the comparison of training Diffusion-QL and our method on four AntMaze tasks as illustrated in Figure~\ref{fig:ql-eql}, maintaining the same number of diffusion steps $T=5$ for both. It is observed that the performance of Diffusion-QL even drops down as the training step increases, while our method is substantially more stable and achieves higher results throughout all the training processes.
We also included a detailed comparison of training and evaluation times for Gaussian and diffusion policies with Q-ensembles in Table~\ref{tab:time}. Increasing $M$ from 2 to 64 almost does not influence the evaluation time. The diffusion step $T$ has more impact on both training and evaluation time which is a common problem in diffusion models. 
\begin{table}[]

\centering
\vspace{-1.3em}
% \scriptsize
\caption{Computational time comparison with different settings on Antmaze-medium-play-v0. 
Training time is for 1 epoch (1000 training steps) and eval time is for 1000 RL steps. 
% (The diffusion step of a Gaussian Model is 1 here because it takes 1 forward pass to get the action from the Gaussian policy.)
}
% \resizebox{1.\linewidth}{!}{
\begin{tabular}{l|cccc}
\toprule
Policy & Diffusion Step $T$ & \# Critics $M$ & Training Time  &  Eval Time \\
\midrule
% Gaussian & 1  & 2  & 5m 35s & 1.45s \\
Gaussian & 1  & 2  & 5m 35s & 1s 450ms \\

Gaussian & 1  & 64 & 7m 20s & 1s 450ms \\
% Gaussian & 1  & 64 & 7m 20s & 1.45s \\
% Gaussian & 10 & 2  & 5m 35s \\
% Gaussian & 10 & 64 & 7m 24s \\
\midrule
Diffusion & 5  & 2  & 9m 30s & 4s 800ms \\
% Diffusion & 5  & 2  & 9m 30s & 4.8 s \\
Diffusion & 5  & 64 & 11m & 4s 800ms\\
% Diffusion & 5  & 64 & 11m & 4.8s\\

Diffusion & 10 & 2  & 12m 23s & 8s \\

Diffusion & 10 & 64 & 13m 55s& 8s \\
\bottomrule
\end{tabular}
\label{tab:time}
% }
\vspace{-2em}
% \end{table}

\end{table}

\section{Related Work}
% \blue{
% VP-SDE comparison, cite DDIM in sampling
% }

\paragraph{Generative Diffusion Models and Mean-reverting SDEs}
Recent advancements have integrated diffusion models \citep{ho2020denoising,song2020denoising,saharia2022photorealistic,rombach2022high} and SDEs \citep{song2020score,luo2023image,welker22speech,richter2023speech} for realistic generative modeling.
The development of Denoising Diffusion Probabilistic Models \citep{ho2020denoising}
showcases the ability of diffusion models to generate high-fidelity images through iterative reverse diffusion processes guided by deep neural networks, achieving state-of-the-art performance in generative tasks. In \citep{luo2023refusion, welker22speech, richter2023speech}, mean-reverting SDEs are applied to speech processing and image restoration tasks. These SDEs, similar to \eqref{eq:t-sde} but with different parameters, ensure our policy adapts across various distributions without bias. The general applicability of our method is demonstrated in 4 D4RL benchmark domains. 
\blue{The comparison between our SDE and \citep{song2020score} are provided in Appendix~\ref{app-sec:vp-sde}.}

\paragraph{Diffusion Models in Offline RL} Diffusion models in offline RL have gained growing attention for their potent modeling capabilities.
In \citet{janner2022diffuser}, diffusion models are introduced as trajectory planners trained with offline datasets for guided sampling, significantly mitigating compounding errors in model-based planning \citep{xiao2019compounding}.
Diffusion models are also used as data synthesizers \citep{chen2023genaug, yu2023scaling}, generating augmented training data to enhance offline RL robustness. 
Additionally, diffusion models approximate behavior policies \citep{wang2022diffusionpolicy, kang2023edp, hansen2023idql}, integrating Q-learning for policy improvement, though this can lead to overly conservative policies.

\paragraph{Entropy Regularization} In online RL, maximum entropy strategies encourage exploration by maximizing rewards while maintaining high entropy \citep{haarnoja2018sac, haarnoja2018soft}. This approach develops diverse skills \citep{eysenbach2018diversity} and adapts to unseen goals \citep{pan2009survey-transfer}. However, its application in offline RL is challenging due to the multi-modal nature of datasets from various policies and expert demonstrations.

\paragraph{Uncertainty Measurement} Balancing exploration and exploitation is crucial when data is limited. 
Online RL methods like bootstrapped DQN \citep{osband2016bootstrap} and Thompson sampling \citep{lattimore2020bandit} estimate uncertainty for exploration guidance. 
In offline RL, handling uncertainty is critical due to the lack of environment interaction. 
Model-based methods like MOPO \citep{yu2020mopo} and MORel \citep{kidambi2020morel} measure and penalize uncertain model dynamics. Similarly, model-free methods like EDAC \citep{an2021edac} and MSG \citep{ghasemipour2022-msg} use Q-network ensembles to obtain pessimistic value estimations for policy guidance.

\section{Conclusion}
\label{sec:conclusion}
In this work, we present an entropy-regularized diffusion policy for offline RL, introducing mean-reverting SDEs as the base framework to provide tractable entropy.
Our theoretical contributions include deriving an approximated entropy for a diffusion model, enabling its integration as an entropy regularization component within the policy loss function.
We also propose an optimal sampling process, ensuring the fast convergence of action generation from diffusion policy. 
Additionally, we enhance our method by incorporating Q-ensembles to handle the data uncertainty. 
Our experimental results show that combining entropy regularization with the LCB approach leads to a more robust policy, achieving state-of-the-art performance across offline RL benchmarks, particularly in AntMaze tasks with sparse rewards and suboptimal trajectories. 

\vspace{-0.5em}
\paragraph{Future Work}
While the proposed method performs well on most D4RL tasks, the diffusion policy requires longer time when executed on compute- and power-constrained devices. Our future work will investigate real-time policy distillation under time and compute constraints to address this challenge.

\section*{Acknowledgements}
This research was financially supported \emph{Kjell och M{\"a}rta Beijer Foundation} and by the project \emph{Deep probabilistic regression -- new models and learning algorithms} (contract number: 2021-04301) as well as contract number 2023-04546, 
funded by the Swedish Research Council. The work was also partially supported by the Wallenberg AI, Autonomous Systems and Software Program (WASP) funded by the Knut and Alice Wallenberg Foundation.
The computations were enabled by the supercomputing resource Berzelius provided by National Supercomputer Centre at Linköping University and the Knut and Alice Wallenberg foundation.

\newpage
{
    \small
    \bibliographystyle{plainnat}
    \bibliography{main_paper}

\begin{thebibliography}{50}
\providecommand{\natexlab}[1]{#1}
\providecommand{\url}[1]{\texttt{#1}}
\expandafter\ifx\csname urlstyle\endcsname\relax
  \providecommand{\doi}[1]{doi: #1}\else
  \providecommand{\doi}{doi: \begingroup \urlstyle{rm}\Url}\fi

\bibitem[An et~al.(2021)An, Moon, Kim, and Song]{an2021edac}
Gaon An, Seungyong Moon, Jang-Hyun Kim, and Hyun~Oh Song.
\newblock Uncertainty-based offline reinforcement learning with diversified q-ensemble.
\newblock \emph{Advances in neural information processing systems}, 34:\penalty0 7436--7447, 2021.

\bibitem[Anderson(1982)]{anderson1982reverse}
Brian~DO Anderson.
\newblock Reverse-time diffusion equation models.
\newblock \emph{Stochastic Processes and their Applications}, 12\penalty0 (3):\penalty0 313--326, 1982.

\bibitem[Bai et~al.(2022)Bai, Wang, Yang, Deng, Garg, Liu, and Wang]{bai2022pessimistic}
Chenjia Bai, Lingxiao Wang, Zhuoran Yang, Zhihong Deng, Animesh Garg, Peng Liu, and Zhaoran Wang.
\newblock Pessimistic bootstrapping for uncertainty-driven offline reinforcement learning.
\newblock \emph{arXiv preprint arXiv:2202.11566}, 2022.

\bibitem[Chen et~al.(2021)Chen, Lu, Rajeswaran, Lee, Grover, Laskin, Abbeel, Srinivas, and Mordatch]{chen2021DT}
Lili Chen, Kevin Lu, Aravind Rajeswaran, Kimin Lee, Aditya Grover, Misha Laskin, Pieter Abbeel, Aravind Srinivas, and Igor Mordatch.
\newblock Decision transformer: Reinforcement learning via sequence modeling.
\newblock \emph{Advances in neural information processing systems}, 34:\penalty0 15084--15097, 2021.

\bibitem[Chen et~al.(2023)Chen, Kiami, Gupta, and Kumar]{chen2023genaug}
Zoey Chen, Sho Kiami, Abhishek Gupta, and Vikash Kumar.
\newblock Genaug: Retargeting behaviors to unseen situations via generative augmentation.
\newblock \emph{arXiv preprint arXiv:2302.06671}, 2023.

\bibitem[Dhariwal and Nichol(2021)]{dhariwal2021diffusion}
Prafulla Dhariwal and Alexander Nichol.
\newblock Diffusion models beat gans on image synthesis.
\newblock \emph{Advances in neural information processing systems}, 34:\penalty0 8780--8794, 2021.

\bibitem[Eysenbach et~al.(2018)Eysenbach, Gupta, Ibarz, and Levine]{eysenbach2018diversity}
Benjamin Eysenbach, Abhishek Gupta, Julian Ibarz, and Sergey Levine.
\newblock Diversity is all you need: Learning skills without a reward function.
\newblock \emph{arXiv preprint arXiv:1802.06070}, 2018.

\bibitem[Fu et~al.(2020)Fu, Kumar, Nachum, Tucker, and Levine]{fu2020d4rl}
Justin Fu, Aviral Kumar, Ofir Nachum, George Tucker, and Sergey Levine.
\newblock D4rl: Datasets for deep data-driven reinforcement learning.
\newblock \emph{arXiv preprint arXiv:2004.07219}, 2020.

\bibitem[Fujimoto and Gu(2021)]{fujimoto2021minimalist}
Scott Fujimoto and Shixiang~Shane Gu.
\newblock A minimalist approach to offline reinforcement learning.
\newblock \emph{Advances in neural information processing systems}, 34:\penalty0 20132--20145, 2021.

\bibitem[Fujimoto et~al.(2019)Fujimoto, Meger, and Precup]{fujimoto2019off}
Scott Fujimoto, David Meger, and Doina Precup.
\newblock Off-policy deep reinforcement learning without exploration.
\newblock In \emph{International conference on machine learning}, pages 2052--2062. PMLR, 2019.

\bibitem[Ghasemipour et~al.(2022)Ghasemipour, Gu, and Nachum]{ghasemipour2022-msg}
Kamyar Ghasemipour, Shixiang~Shane Gu, and Ofir Nachum.
\newblock Why so pessimistic? estimating uncertainties for offline rl through ensembles, and why their independence matters.
\newblock \emph{Advances in Neural Information Processing Systems}, 35:\penalty0 18267--18281, 2022.

\bibitem[Gillespie(1996)]{gillespie1996exact}
Daniel~T Gillespie.
\newblock Exact numerical simulation of the ornstein-uhlenbeck process and its integral.
\newblock \emph{Physical review E}, 54\penalty0 (2):\penalty0 2084, 1996.

\bibitem[Haarnoja et~al.(2017)Haarnoja, Tang, Abbeel, and Levine]{haarnoja2017reinforcement}
Tuomas Haarnoja, Haoran Tang, Pieter Abbeel, and Sergey Levine.
\newblock Reinforcement learning with deep energy-based policies.
\newblock In \emph{International conference on machine learning}, pages 1352--1361. PMLR, 2017.

\bibitem[Haarnoja et~al.(2018{\natexlab{a}})Haarnoja, Zhou, Abbeel, and Levine]{haarnoja2018sac}
Tuomas Haarnoja, Aurick Zhou, Pieter Abbeel, and Sergey Levine.
\newblock Soft actor-critic: Off-policy maximum entropy deep reinforcement learning with a stochastic actor.
\newblock In \emph{International conference on machine learning}, pages 1861--1870. PMLR, 2018{\natexlab{a}}.

\bibitem[Haarnoja et~al.(2018{\natexlab{b}})Haarnoja, Zhou, Abbeel, and Levine]{haarnoja2018soft}
Tuomas Haarnoja, Aurick Zhou, Pieter Abbeel, and Sergey Levine.
\newblock Soft actor-critic: Off-policy maximum entropy deep reinforcement learning with a stochastic actor.
\newblock In \emph{International conference on machine learning}, pages 1861--1870. PMLR, 2018{\natexlab{b}}.

\bibitem[Haarnoja et~al.(2018{\natexlab{c}})Haarnoja, Zhou, Hartikainen, Tucker, Ha, Tan, Kumar, Zhu, Gupta, Abbeel, et~al.]{haarnoja2018soft2}
Tuomas Haarnoja, Aurick Zhou, Kristian Hartikainen, George Tucker, Sehoon Ha, Jie Tan, Vikash Kumar, Henry Zhu, Abhishek Gupta, Pieter Abbeel, et~al.
\newblock Soft actor-critic algorithms and applications.
\newblock \emph{arXiv preprint arXiv:1812.05905}, 2018{\natexlab{c}}.

\bibitem[Hansen-Estruch et~al.(2023)Hansen-Estruch, Kostrikov, Janner, Kuba, and Levine]{hansen2023idql}
Philippe Hansen-Estruch, Ilya Kostrikov, Michael Janner, Jakub~Grudzien Kuba, and Sergey Levine.
\newblock Idql: Implicit q-learning as an actor-critic method with diffusion policies.
\newblock \emph{arXiv preprint arXiv:2304.10573}, 2023.

\bibitem[Ho et~al.(2020)Ho, Jain, and Abbeel]{ho2020denoising}
Jonathan Ho, Ajay Jain, and Pieter Abbeel.
\newblock Denoising diffusion probabilistic models.
\newblock \emph{Advances in neural information processing systems}, 33:\penalty0 6840--6851, 2020.

\bibitem[Janner et~al.(2022)Janner, Du, Tenenbaum, and Levine]{janner2022diffuser}
Michael Janner, Yilun Du, Joshua~B Tenenbaum, and Sergey Levine.
\newblock Planning with diffusion for flexible behavior synthesis.
\newblock \emph{arXiv preprint arXiv:2205.09991}, 2022.

\bibitem[Jin et~al.(2021)Jin, Yang, and Wang]{jin2021pessimism}
Ying Jin, Zhuoran Yang, and Zhaoran Wang.
\newblock Is pessimism provably efficient for offline rl?
\newblock In \emph{International Conference on Machine Learning}, pages 5084--5096. PMLR, 2021.

\bibitem[Kang et~al.(2023{\natexlab{a}})Kang, Ma, Du, Pang, and Yan]{kang2023edp}
Bingyi Kang, Xiao Ma, Chao Du, Tianyu Pang, and Shuicheng Yan.
\newblock Efficient diffusion policies for offline reinforcement learning.
\newblock \emph{arXiv preprint arXiv:2305.20081}, 2023{\natexlab{a}}.

\bibitem[Kang et~al.(2023{\natexlab{b}})Kang, Ma, Du, Pang, and Yan]{kang2023efficient}
Bingyi Kang, Xiao Ma, Chao Du, Tianyu Pang, and Shuicheng Yan.
\newblock Efficient diffusion policies for offline reinforcement learning.
\newblock \emph{arXiv preprint arXiv:2305.20081}, 2023{\natexlab{b}}.

\bibitem[Kidambi et~al.(2020)Kidambi, Rajeswaran, Netrapalli, and Joachims]{kidambi2020morel}
Rahul Kidambi, Aravind Rajeswaran, Praneeth Netrapalli, and Thorsten Joachims.
\newblock Morel: Model-based offline reinforcement learning.
\newblock \emph{Advances in neural information processing systems}, 33:\penalty0 21810--21823, 2020.

\bibitem[Kingma and Ba(2014)]{kingma2014adam}
Diederik~P Kingma and Jimmy Ba.
\newblock Adam: A method for stochastic optimization.
\newblock \emph{arXiv preprint arXiv:1412.6980}, 2014.

\bibitem[Kingma and Welling(2013)]{kingma2013auto}
Diederik~P Kingma and Max Welling.
\newblock Auto-encoding variational bayes.
\newblock \emph{arXiv preprint arXiv:1312.6114}, 2013.

\bibitem[Kostrikov et~al.(2021)Kostrikov, Nair, and Levine]{kostrikov2021IQL}
Ilya Kostrikov, Ashvin Nair, and Sergey Levine.
\newblock Offline reinforcement learning with implicit q-learning.
\newblock \emph{arXiv preprint arXiv:2110.06169}, 2021.

\bibitem[Kumar et~al.(2019)Kumar, Fu, Soh, Tucker, and Levine]{kumar2019stabilizing}
Aviral Kumar, Justin Fu, Matthew Soh, George Tucker, and Sergey Levine.
\newblock Stabilizing off-policy q-learning via bootstrapping error reduction.
\newblock \emph{Advances in Neural Information Processing Systems}, 32, 2019.

\bibitem[Kumar et~al.(2020)Kumar, Zhou, Tucker, and Levine]{kumar2020conservative}
Aviral Kumar, Aurick Zhou, George Tucker, and Sergey Levine.
\newblock Conservative q-learning for offline reinforcement learning.
\newblock \emph{Advances in Neural Information Processing Systems}, 33:\penalty0 1179--1191, 2020.

\bibitem[Lange et~al.(2012)Lange, Gabel, and Riedmiller]{lange2012batchrl}
Sascha Lange, Thomas Gabel, and Martin Riedmiller.
\newblock Batch reinforcement learning.
\newblock In \emph{Reinforcement learning: State-of-the-art}, pages 45--73. Springer, 2012.

\bibitem[Lattimore and Szepesv{\'a}ri(2020)]{lattimore2020bandit}
Tor Lattimore and Csaba Szepesv{\'a}ri.
\newblock \emph{Bandit algorithms}.
\newblock Cambridge University Press, 2020.

\bibitem[Levine et~al.(2020)Levine, Kumar, Tucker, and Fu]{levine2020offline}
Sergey Levine, Aviral Kumar, George Tucker, and Justin Fu.
\newblock Offline reinforcement learning: Tutorial, review, and perspectives on open problems.
\newblock \emph{arXiv preprint arXiv:2005.01643}, 2020.

\bibitem[Luo et~al.(2023{\natexlab{a}})Luo, Gustafsson, Zhao, Sj{\"o}lund, and Sch{\"o}n]{luo2023image}
Ziwei Luo, Fredrik~K Gustafsson, Zheng Zhao, Jens Sj{\"o}lund, and Thomas~B Sch{\"o}n.
\newblock Image restoration with mean-reverting stochastic differential equations.
\newblock \emph{International Conference on Machine Learning}, 2023{\natexlab{a}}.

\bibitem[Luo et~al.(2023{\natexlab{b}})Luo, Gustafsson, Zhao, Sj{\"o}lund, and Sch{\"o}n]{luo2023refusion}
Ziwei Luo, Fredrik~K Gustafsson, Zheng Zhao, Jens Sj{\"o}lund, and Thomas~B Sch{\"o}n.
\newblock Refusion: Enabling large-size realistic image restoration with latent-space diffusion models.
\newblock In \emph{Proceedings of the IEEE/CVF conference on computer vision and pattern recognition}, pages 1680--1691, 2023{\natexlab{b}}.

\bibitem[Misra(2019)]{misra2019mish}
Diganta Misra.
\newblock Mish: A self regularized non-monotonic activation function.
\newblock \emph{arXiv preprint arXiv:1908.08681}, 2019.

\bibitem[Mnih et~al.(2016)Mnih, Badia, Mirza, Graves, Lillicrap, Harley, Silver, and Kavukcuoglu]{mnih2016asynchronous}
Volodymyr Mnih, Adria~Puigdomenech Badia, Mehdi Mirza, Alex Graves, Timothy Lillicrap, Tim Harley, David Silver, and Koray Kavukcuoglu.
\newblock Asynchronous methods for deep reinforcement learning.
\newblock In \emph{International conference on machine learning}, pages 1928--1937. PMLR, 2016.

\bibitem[Nichol and Dhariwal(2021)]{nichol2021improved}
Alexander~Quinn Nichol and Prafulla Dhariwal.
\newblock Improved denoising diffusion probabilistic models.
\newblock In \emph{International Conference on Machine Learning}, pages 8162--8171. PMLR, 2021.

\bibitem[Osband et~al.(2016)Osband, Blundell, Pritzel, and Van~Roy]{osband2016bootstrap}
Ian Osband, Charles Blundell, Alexander Pritzel, and Benjamin Van~Roy.
\newblock Deep exploration via bootstrapped {DQN}.
\newblock \emph{Advances in neural information processing systems}, 29, 2016.

\bibitem[Pan and Yang(2009)]{pan2009survey-transfer}
Sinno~Jialin Pan and Qiang Yang.
\newblock A survey on transfer learning.
\newblock \emph{IEEE Transactions on knowledge and data engineering}, 22\penalty0 (10):\penalty0 1345--1359, 2009.

\bibitem[Richter et~al.(2023)Richter, Welker, Lemercier, Lay, and Gerkmann]{richter2023speech}
Julius Richter, Simon Welker, Jean-Marie Lemercier, Bunlong Lay, and Timo Gerkmann.
\newblock Speech enhancement and dereverberation with diffusion-based generative models.
\newblock \emph{IEEE/ACM Transactions on Audio, Speech, and Language Processing}, 2023.

\bibitem[Rombach et~al.(2022)Rombach, Blattmann, Lorenz, Esser, and Ommer]{rombach2022high}
Robin Rombach, Andreas Blattmann, Dominik Lorenz, Patrick Esser, and Bj{\"o}rn Ommer.
\newblock High-resolution image synthesis with latent diffusion models.
\newblock In \emph{Proceedings of the IEEE/CVF conference on computer vision and pattern recognition}, pages 10684--10695, 2022.

\bibitem[Saharia et~al.(2022)Saharia, Chan, Saxena, Li, Whang, Denton, Ghasemipour, Gontijo~Lopes, Karagol~Ayan, Salimans, et~al.]{saharia2022photorealistic}
Chitwan Saharia, William Chan, Saurabh Saxena, Lala Li, Jay Whang, Emily~L Denton, Kamyar Ghasemipour, Raphael Gontijo~Lopes, Burcu Karagol~Ayan, Tim Salimans, et~al.
\newblock Photorealistic text-to-image diffusion models with deep language understanding.
\newblock \emph{Advances in neural information processing systems}, 35:\penalty0 36479--36494, 2022.

\bibitem[Song et~al.(2020{\natexlab{a}})Song, Meng, and Ermon]{song2020denoising}
Jiaming Song, Chenlin Meng, and Stefano Ermon.
\newblock Denoising diffusion implicit models.
\newblock \emph{arXiv preprint arXiv:2010.02502}, 2020{\natexlab{a}}.

\bibitem[Song et~al.(2020{\natexlab{b}})Song, Sohl-Dickstein, Kingma, Kumar, Ermon, and Poole]{song2020score}
Yang Song, Jascha Sohl-Dickstein, Diederik~P Kingma, Abhishek Kumar, Stefano Ermon, and Ben Poole.
\newblock Score-based generative modeling through stochastic differential equations.
\newblock \emph{arXiv preprint arXiv:2011.13456}, 2020{\natexlab{b}}.

\bibitem[Wang et~al.(2022)Wang, Hunt, and Zhou]{wang2022diffusionpolicy}
Zhendong Wang, Jonathan~J Hunt, and Mingyuan Zhou.
\newblock Diffusion policies as an expressive policy class for offline reinforcement learning.
\newblock \emph{arXiv preprint arXiv:2208.06193}, 2022.

\bibitem[Welker et~al.(2022)Welker, Richter, and Gerkmann]{welker22speech}
Simon Welker, Julius Richter, and Timo Gerkmann.
\newblock Speech enhancement with score-based generative models in the complex {STFT} domain.
\newblock In \emph{Proc. Interspeech 2022}, pages 2928--2932, 2022.
\newblock \doi{10.21437/Interspeech.2022-10653}.

\bibitem[Xiao et~al.(2019)Xiao, Wu, Ma, Schuurmans, and M{\"u}ller]{xiao2019compounding}
Chenjun Xiao, Yifan Wu, Chen Ma, Dale Schuurmans, and Martin M{\"u}ller.
\newblock Learning to combat compounding-error in model-based reinforcement learning.
\newblock \emph{arXiv preprint arXiv:1912.11206}, 2019.

\bibitem[Yu et~al.(2020)Yu, Thomas, Yu, Ermon, Zou, Levine, Finn, and Ma]{yu2020mopo}
Tianhe Yu, Garrett Thomas, Lantao Yu, Stefano Ermon, James~Y Zou, Sergey Levine, Chelsea Finn, and Tengyu Ma.
\newblock Mopo: Model-based offline policy optimization.
\newblock \emph{Advances in Neural Information Processing Systems}, 33:\penalty0 14129--14142, 2020.

\bibitem[Yu et~al.(2023)Yu, Xiao, Stone, Tompson, Brohan, Wang, Singh, Tan, Peralta, Ichter, et~al.]{yu2023scaling}
Tianhe Yu, Ted Xiao, Austin Stone, Jonathan Tompson, Anthony Brohan, Su~Wang, Jaspiar Singh, Clayton Tan, Jodilyn Peralta, Brian Ichter, et~al.
\newblock Scaling robot learning with semantically imagined experience.
\newblock \emph{arXiv preprint arXiv:2302.11550}, 2023.

\bibitem[Zhu et~al.(2023)Zhu, Zhao, He, Zhong, Zhang, Yu, and Zhang]{zhu2023diffusion}
Zhengbang Zhu, Hanye Zhao, Haoran He, Yichao Zhong, Shenyu Zhang, Yong Yu, and Weinan Zhang.
\newblock Diffusion models for reinforcement learning: A survey.
\newblock \emph{arXiv preprint arXiv:2311.01223}, 2023.

\bibitem[Ziebart(2010)]{ziebart2010modeling}
Brian~D Ziebart.
\newblock \emph{Modeling purposeful adaptive behavior with the principle of maximum causal entropy}.
\newblock Carnegie Mellon University, 2010.

\end{thebibliography}
}

\newpage
\appendix

\section*{Appendix}
\section{Proof}
\label{app-sec:proof}

\subsection{Solution to the Forward SDE}
\label{prf:SDE_solution}

Given the forward Stochastic Differential Equation (SDE) represented by
\begin{equation}
    \diff \bm{a} = -\theta_t \bm{a} \diff t + \sigma_t \diff \bm{w}, \quad \bm{a}^0 \sim p_0(\bm{a}),
    \label{app-eq:forward_sde}
\end{equation}
where $\theta_t$ and $\sigma_t$ are time-dependent positive functions, and $\bm{w}$ denotes a standard Wiener process. We consider the special case where $\sigma_t^2 = 2\theta_t$ for all $t$. The solution for the transition probability from time $\tau$ to $t$ ($\tau < t$) is given by

\begin{equation}
    p(\bm{a}^t | \bm{a}^\tau) = \mathcal{N}\left(\bm{a}^t | \bm{a}^\tau \expp^{-\bar{\theta}_{\tau:t}}, (1 - \expp^{-2 \bar{\theta}_{\tau:t}}) \bm{I}\right).
    \label{app-eq:sde_solution}
\end{equation}

\begin{proof}
The proof is in general similar to that in IR-SDE~\cite{luo2023image}. To solve Equation \eqref{app-eq:forward_sde},  we introduce the transformation 
\begin{equation}
    \psi(\bm{a}, t) = \bm{a} \expp^{\bar{\theta}_t},
    \label{eq:psi_transformation}
\end{equation}
and apply It\^{o}'s formula to obtain
\begin{equation}
    \diff \psi(\bm{a}, t) = \sigma_t \expp^{\bar{\theta}_t} \diff \bm{w}(t).
    \label{eq:ito_application}
\end{equation}
Integrating from $\tau$ to $t$, we get
\begin{equation}
    \psi(\bm{a}^t, t) - \psi(\bm{a}^\tau, \tau) = \int_{\tau}^t \sigma_z \expp^{\bar{\theta}_z} \diff \bm{w}(z),
    \label{eq:psi_integration}
\end{equation}
we can analytically compute the two integrals as ${{\theta}_t}$ and $\bm{\sigma}_t$ are scalars and then obtain
\begin{equation}
    \bm{a}(t) \expp^{\bar{\theta}_t} - \bm{a}^\tau \expp^{\bar{\theta}_\tau} = \int_{\tau}^t \sigma_z \expp^{\bar{\theta}_z} \diff \bm{w}(z).
    \label{eq:simplified_psi}
\end{equation}
Rearranging terms and dividing by $\expp^{\bar{\theta}_t}$, we obtain
\begin{equation}
    \bm{a}(t) = \bm{a}(\tau) \expp^{-\bar{\theta}_{\tau:t}} + \int^t_\tau \sigma_z \expp^{-\bar{\theta}_{z:t}} \diff \bm{w}(z).
    \label{eq:rearranged_solution}
\end{equation}
The integral term is actually a Gaussian random variable with mean zero and variance
\begin{equation}
    \int_{\tau}^t \sigma_z^2 \expp^{-2\bar{\theta}_{z:t}} \diff z = \lambda^2 (1 - \expp^{-2 \bar{\theta}_{\tau:t}}),
    \label{eq:variance_integral}
\end{equation}
under the condition $\sigma_t^2 = 2\theta_t$. Thus, the transition probability is
\begin{equation}
    p(\bm{a}^t | \bm{a}^\tau) = \mathcal{N}(\bm{a}^t | \bm{a}^\tau \expp^{-\bar{\theta}_{\tau:t}},  (1 - \expp^{-2 \bar{\theta}_{\tau:t}})\bm{I}).
    \label{eq:final_transition_probability}
\end{equation}
This completes the proof.
\end{proof}

\paragraph{Loss function}
% \subsection{Loss function}
From \eqref{app-eq:sde_solution}, the marginal distribution of $p(\bm{a}(t))$ can be written as
\begin{equation}
\begin{split}
    p(\bm{a}(t)) 
    &= p(\bm{a}(t) \cond \bm{a}(0)) \\ 
    &= \mathcal{N}(\bm{a}(t) \cond \bm{a}(0) \expp^{-\bar{\theta}_{t}}, (1 - \expp^{-2 \, \bar{\theta}_{t}}) \mathbf{I}).
\end{split}
\label{eq:margin_p}
\end{equation}
where we substitute $\bar{\theta}_{0:t}$ with $\bar{\theta}_{t}$ for clear notation.

During training, the initial diffusion state ${\bm a}^0$ is given and thus we can obtain the ground truth score $\nabla_{{\bm a}}\log p_t({\bm a})$ based on the marginal distribution:
\begin{equation}
    \nabla_{{\bm a}}\log p_t({{\bm a}} \cond {{\bm a}}_0) = - \frac{{{\bm a}}_t - {\bm a}^0 \expp^{-\bar{\theta}_{t}}}{1 - \expp^{-2 \, \bar{\theta}_{t}}},
    \label{eq:score}
\end{equation}
which can be approximated using a neural network and optimized with score-matching loss. Moreover, the marginal distribution \eqref{eq:margin_p} gives the reparameterization of the state:
\begin{equation}
    {\bm a}^t = {\bm a}^0\expp^{-\bar{\theta}_{t}} + \sqrt{1 - \expp^{-2\bar{\theta}_{t}}} \cdot \bm{\epsilon}_t,
    \label{app-eq:reparameterize_xt}
\end{equation}
where $\bm{\epsilon}_t$ is a standard Gaussian noise $\bm{\epsilon}_t \sim \mathcal{N}(0, \bm{I})$. By substituting \eqref{app-eq:reparameterize_xt} into \eqref{eq:score}, the score function can be re-written in terms of the noise as
\begin{equation}
    \nabla_{{\bm a}}\log p_t({{\bm a}} \cond {{\bm a}}_0) = - \frac{\bm{\epsilon}_t}{\sqrt{1 - \expp^{-2 \, \bar{\theta}_{t}}}}.
    \label{eq:noise-score}
\end{equation}
Then we follow the practical settings in diffusion models~\citep{ho2020denoising,dhariwal2021diffusion} to estimate the noise with a time-dependent neural network ${\bm{\epsilon}}_\phi$ and optimize it with a simplified noise matching loss:
\begin{equation}
    L(\phi) \coloneqq \mathbb{E}_{t \in [0, T]} \Big[ \big\lVert {\bm{\epsilon}}_\phi({\bm a}^0\expp^{-\bar{\theta}_{t}} + \sqrt{1 - \expp^{-2\bar{\theta}_{t}}} \cdot \bm{\epsilon}_t, t) - \bm{\epsilon}_t) \bigr\rVert\Big],
    \label{app-eq:noise_objective}
\end{equation}
where $t$ is a randomly sampled timestep and $\{{{\bm a}}_t\}_{t=0}^T$ denotes the discretization of the diffusion process. And this loss \eqref{app-eq:noise_objective} is the same as \eqref{eq:noise_objective} in the main paper.

\subsection{Sampling from the Posterior}
\label{prf:posterior}

\textbf{Proposition 3.1.}
% \begin{proposition}
    \textit{Given an initial variable ${\bm a}^0$, for any diffusion state ${\bm a}^t$ at time $t \in [1,T]$, the posterior of the mean-reverting SDE \eqref{eq:t-sde} conditioned on ${\bm a}^0$ is} 
    \begin{equation}
        p({\bm a}^{t-1} \cond {\bm a}^t, {\bm a}^0) = \mathcal{N}({\bm a}^{t-1} \cond \tilde{\mu}_t({\bm a}^t, \, {\bm a}^0), \; \tilde{\beta}_t \mathbf{I}),
    % \label{eq:posterior}
    \end{equation}
    \textit{which is a Gaussian with mean and variance given by:}
    \begin{equation}
    \begin{split}
        \tilde{\mu}_t({\bm a}^t, {\bm a}^0) &\coloneqq \frac{1 - \expp^{-2\bar{\theta}_{t-1}}}{1 - \expp^{-2\bar{\theta}_{t}}}{\expp^{-\theta_t}} {\bm a}^t + \frac{1 - \expp^{-2\theta_t}}{1 - \expp^{-2\bar{\theta}_{t}}}{\expp^{-\bar{\theta}_{t-1}}} {\bm a}^0 \\[2mm]
        \mathrm{and} \quad \tilde{\beta}_t &\coloneqq \frac{(1 - {\expp^{-2\bar{\theta}_{t-1}}})(1 - {\expp^{-2\theta^{'}_t}})}{1 - \expp^{-2\bar{\theta}_{t}}},
        % \label{eq:posterior_mu_var}
    \end{split}
    \end{equation}
    \textit{where $\theta_i^{'} \coloneqq \int_{i-1}^i \theta_t dt$ and $\bar{\theta}_{t}$ is to substitute  $\bar{\theta}_{0:t}$ for clear notation.}
    %
% \end{proposition}

\begin{proof}
The posterior of SDE can be derived from Bayes' rule,
\begin{equation}
    p({\bm a}^{{t-1}} \mid {\bm a}^{t}, {\bm a}^0) = \frac{p({\bm a}^{t} \mid {\bm a}^{{t-1}}, {\bm a}^0) p({\bm a}^{{t-1}} \mid {\bm a}^0)}{p({\bm a}^{t} \mid {\bm a}^0)}.
    \label{app-eq:cond_prob_bayes}
\end{equation} Recall that the transition distribution $p({\bm a}^{t} \cond {\bm a}^{t-1})$ and $p({\bm a}^{t} \cond {\bm a}_{0})$ can be known with the solution to the forward SDE. Since all the distributions are Gaussian, the posterior will also be a Gaussian. 
\small
\begin{align}
\begin{aligned}
&p({\bm a}^{{t-1}} \mid {\bm a}^{t}, {\bm a}^0) \\
& \propto \exp \left(
    -\frac{1}{2}\left(
      \frac{ (\bm a^t- \bm a^{t-1}\expp^{-\theta^{'}_t} )^2}{ 1 - \expp^{-2 \theta^{'}_t}} +
    \frac{(\bm a^{t-1} - \bm a^0\expp^{-\bar{\theta}_{t-1}} )^2}{1 - \expp^{-2 \bar{\theta}_{t-1}}} -
    \frac{(\bm a^{t} - \bm a^0\expp^{-\bar{\theta}_{t}} )^2}{1 - \expp^{-2 \bar{\theta}_{t}}}
    \right) 
\right)\\
&= \exp \left(
-\frac{1}{2}\left(
   (\frac{\expp^{-2\theta^{'}_t}}{1 - \expp^{-2 \theta^{'}_t} }
    + \frac{1}{1 - \expp^{-2 \bar{\theta}_{t-1}}}
   ) (\bm a^{t-1})^2 
   - (\frac{2 \expp^{-\theta^{'}_t} }{1 - \expp^{-2 \theta^{'}_t} }\bm a^{t} + \frac{2 \expp^{-\bar{\theta}_{t-1}} }{1 - \expp^{-2 \bar{\theta}_{t-1}}}\bm a^0) \bm a^{t-1} 
   + C(\bm a^{t}, \bm a^{0})
\right)
\right)
\end{aligned}
\end{align}
\normalsize
where $C(\bm a^{t}, \bm a^{0})$ is some function not involving $\bm (a^{t-1})^2$.
With the standard Gaussian density function, the mean and the variance can be computed:
\begin{equation}
    \begin{split}
        \tilde{\mu}_t({\bm a}^t, {\bm a}^0) &\coloneqq \frac{1 - \expp^{-2\bar{\theta}_{t-1}}}{1 - \expp^{-2\bar{\theta}_{t}}}{\expp^{-\theta^{'}_t}} {\bm a}^t + \frac{1 - \expp^{-2\theta^{'}_t}}{1 - \expp^{-2\bar{\theta}_{t}}}{\expp^{-\bar{\theta}_{t-1}}} {\bm a}^0 \\[2mm]
        \mathrm{and} \quad \tilde{\beta}_t &\coloneqq \frac{(1 - {\expp^{-2\bar{\theta}_{t-1}}})(1 - {\expp^{-2\theta^{'}_t}})}{1 - \expp^{-2\bar{\theta}_{t}}}.
    \end{split}
    \end{equation}
Thus we complete the proof.
\end{proof}

\subsection{Optimal Reverse Path}
\label{prf:optimal_reverse}

In addition, we can prove the distribution mean $\tilde{\mu}_t({\bm a}^t, {\bm a}^0)$ is the optimal reverse path from ${\bm a}^t$ to ${\bm a}^{t-1}$.
\label{prf:optimal_reverse_path}
\begin{proof}

As stated in Proposition~3.1, the posterior is a Gaussian distribution and can be derived by Bayes' rule. Thus it is natural to find the optimal reverse path by minimizing the negative log-likelihood according to 
\begin{equation}
\begin{split}
    {\bm a}^{t-1}_{*} = \arg\min_{{\bm a}^{t-1}} \Bigl[ -\log p \bigl({\bm a}^{t-1} \mid {\bm a}^t, {\bm a}^0 \bigr) \Bigr].
    \label{app-eq:cond_prob_nll}
\end{split}
\end{equation}

From \eqref{app-eq:cond_prob_bayes}, we have
\begin{equation}
\begin{split}
    -\log p \bigl({\bm a}^{t-1} \mid {\bm a}^t, {\bm a}^0 \bigr) 
    & \propto -\log p\bigl({\bm a}^{i} \mid {\bm a}^{t-1}, {\bm a}^0\bigr) - \log p\bigl({\bm a}^{t-1} \mid {\bm a}^0\bigr)
\end{split}
\label{app-eq:cond_prob_bayes_log}
\end{equation}
Then we can directly solve~\eqref{app-eq:cond_prob_nll} by computing the gradient of the negative log-likelihood and setting it to $0$:

\begin{equation}
\begin{split}
     \nabla_{{\bm a}^{t-1}_{*}} \left\{-\log p \bigl({\bm a}^{t-1}_{*} \mid {\bm a}^t, {\bm a}^0 \bigr)\right\} 
     & \propto - \nabla_{{\bm a}^{t-1}_{*}}\log p\bigl({\bm a}^{i} \mid {\bm a}^{t-1}_{*}, {\bm a}^0\bigr) - \nabla_{{\bm a}^{t-1}_{*}} \log p\bigl({\bm a}^{t-1}_{*} \mid {\bm a}^0\bigr) \\[.6em]
     & = - \frac{\expp^{-{\theta}_t^{'}} ({\bm a}^t  - {\bm a}^{t-1}_{*} \expp^{-{\theta}_t^{'}})}{1 - \expp^{-2 \, {\theta}_t^{'}}} + \frac{{\bm a}^{t-1}_{*}  - {\bm a}^0 \expp^{-\bar{\theta}_{t-1}}}{1 - \expp^{-2 \, \bar{\theta}_{t-1}}} \\[.6em]
     & = \frac{{\bm a}^{t-1}_{*} \expp^{-2{\theta}_t^{'}}}{1 - \expp^{-2 \, {\theta}_t^{'}}} + \frac{{\bm a}^{t-1}_{*} }{1 - \expp^{-2 \, \bar{\theta}_{t-1}}} - \frac{{\bm a}^{i} \expp^{-{\theta}_t^{'}}}{1 - \expp^{-2 \, {\theta}_t^{'}}} - \frac{{\bm a}_{0} \expp^{-\bar{\theta}_{t-1}}}{1 - \expp^{-2 \, \bar{\theta}_{t-1}}} \\[.6em]
     & = \frac{{\bm a}^{t-1}_{*} (1 - \expp^{-2 \, \bar{\theta}_{i}})}{(1 - \expp^{-2 \, {\theta}_t^{'}})(1 - \expp^{-2 \, \bar{\theta}_{t-1}})} - \frac{{\bm a}^{i} \expp^{-{\theta}_t^{'}}}{1 - \expp^{-2 \, {\theta}_t^{'}}} - \frac{{\bm a}_{0} \expp^{-\bar{\theta}_{t-1}}}{1 - \expp^{-2 \, \bar{\theta}_{t-1}}} = 0.
    \label{app-eq:solve_nll}
\end{split}
\end{equation}
Since \eqref{app-eq:solve_nll} is linear, we get
\begin{equation}
    \begin{split}
        {\bm a}^{t-1}_{*} = \frac{1 - \expp^{-2 \, \bar{\theta}_{t-1}}}{1 - \expp^{-2 \, \bar{\theta}_t}} \expp^{-\theta_t^{'}} {\bm a}^t  + \frac{1 - \expp^{-2 \, \theta_t^{'}}}{1 - \expp^{-2 \, \bar{\theta}_t}} \expp^{-\bar{\theta}_{t-1}} {\bm a}^0.
    \end{split}
\end{equation}
\end{proof}
This completes the proof. Note that the second-order derivative is a positive constant, and thus ${\bm a}^{t-1}_{*}$ is the optimal point. And we find that this optimal reverse path is the same as our posterior distribution mean as shown in Proposition~\ref{prop:posterior}.

\subsection{Entropy Approximation}
\label{prf:entropy_approximation}

Proposition~\ref{prop:posterior} further shows that the conditional posterior from ${\bm a}_i^1$ to ${\bm a}_i^0$ is Gaussian, meaning that 
\begin{equation}
    -\log\pi_\phi({\bm a}_i^0 \cond {\bm s}_i) = -\log\pi_\phi({\bm a}_i^1 \cond {\bm s}_i) + \mathcal{C},
    \label{app-eq:a0_propto_a1}
\end{equation}
where $\mathcal{C}$ is a constant and ${\bm a}_i^1$ can be approximated using 

\begin{equation}
    p(\bm{a}^t \cond \bm{a}^\tau) = 
    \mathcal{N}(\bm{a}^t \cond \bm{a}^\tau \expp^{-\bar{\theta}_{\tau:t}}, (1 - \expp^{-2 \, \bar{\theta}_{\tau:t}}) \bm{I}),
    \label{app-eq:sde-solution}
\end{equation}

\begin{proof}
Let's consider \textbf{sequentially sampled action states} $a_i^0, a_i^1$ from our optimal sampling strategy. Then we have
\begin{equation}
    \pi(a_i^0 | s_i) = \pi(a_i^0 | a_i^1, {\hat a}_i^0) \cdot \pi(a_i^1 | s_i),
\end{equation}
where ${\hat a}_i^0$ is the approximated action's initial state from Eq. (8). Proposition 3.1 shows that the conditional posterior from $a_i^1$ to $a_i^0$ is a Gaussian with \textbf{certain mean and variance}, meaning that the term $\pi(a_i^0 | a_i^1, {\hat a}_i^0)$ is a computable constant and thus we can write
\begin{equation}
    -\log\pi(a_i^0 | s_i) = -\log\pi(a_i^0 | a_i^1, {\hat a}_i^0) -\log\pi(a_i^1 | s_i) = -\log\pi(a_i^1 | s_i) + \mathcal{C}.
\end{equation}
\end{proof}
\subsection{Visualization}
\label{app-sec:visualization}
\begin{figure}[ht]
    \centering
        \includegraphics[width=0.85\linewidth]{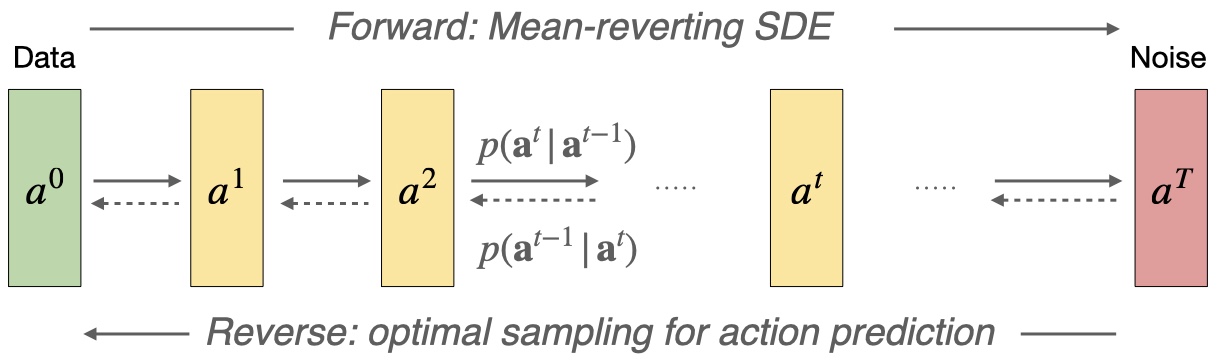}
        \caption{Visualization of the workings of the mean-reverting SDE for action prediction. The SDE models the degradation process from the action from the dataset to a noise. By guiding the policy with corresponding reverse-time SDE and the LCB of Q, a new action is generated conditioned on the RL state.}
    \label{fig:eql-visual}
\end{figure}

For a more intuitive explanation of our approach, Figure~\ref{fig:eql-visual} outlines the forward and reverse processes of the mean-reverting SDE used for action prediction.

\subsection{Comparison to VP SDE}
\label{app-sec:vp-sde}
\blue{
Our mean-reverting SDE is derived from the well-known Ornstein-Uhlenbeck (OU) process \cite{gillespie1996exact} which has the following form: $$ \mathrm{d} x = \theta (\mu - x) \mathrm{d} t + \sigma \mathrm{d}w. $$ As $t \to \infty$, its marginal distribution $p_t(x)$ converges to a stationary Gaussian with mean value $\mu$, which explains the name: “mean-reverting”. We assume that there is no prior knowledge of the actions and thus set $\mu = 0$ to generate actions from standard Gaussian noise. Then, with $\mu = 0$, the mean-reverting SDE has the same form as VP SDE. However, in \cite{song2020score}, no solution of the continuous-time SDE was given. The authors start from perturbing data with multiple noise scales and generalize this idea with an infinite number of noise scales which makes the perturbed data distributions evolve according to an SDE. They keep using the solution of DDPM while we use Itô's formula to solve the continuous-time SDE. Compared to the original VP SDE, our mean-reverting SDE is analytically tractable, see~\eqref{eq:sde-solution} and thus its score $\nabla_{x} \log p_t(x)$ is easier to learn. More importantly, the solution of the mean-reverting SDE can be used for entropy approximation.
}

\section{Additional Experiments Details}
\label{appedix:experiments}

\subsection{More experiments on proposed optimal sampling with different sample step}
\label{app-sec:optimal_sampling}
We added the figures of data generation with fewer steps ($T=1$ and $T=2$) for the toy task in Section 3.1. The results show that the optimal sampling strategy significantly outperforms the reverse-time SDE in all steps, further demonstrating the efficiency and effectiveness of our method.
\begin{figure}[ht]
    \includegraphics[width=.99\linewidth]{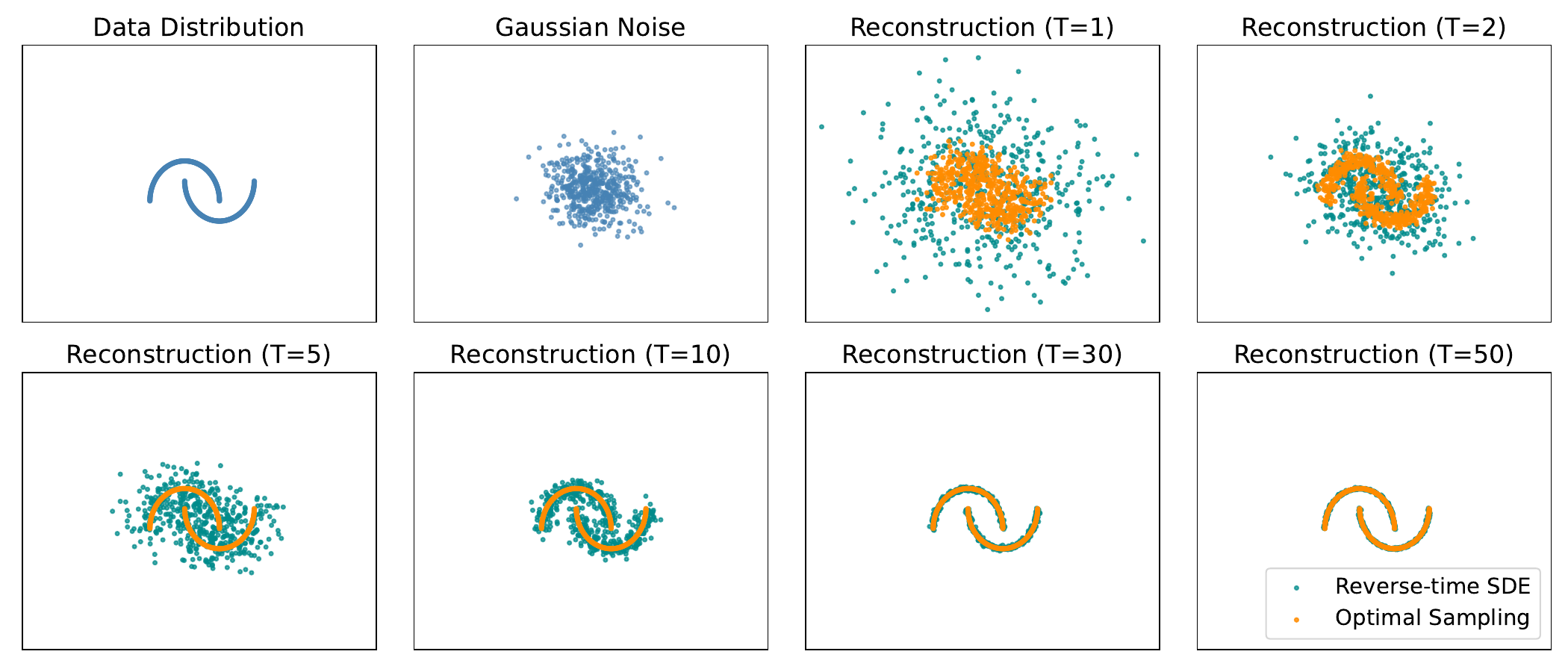}
        \caption{The proposed optimal sampling with different sample steps.}
        \label{app-fig:optimal_sampling}
\end{figure}

\subsection{Hyperparameters}
\label{appendix:hyper}
As stated in Section~\ref{sec:exp:hyper}, we keep our key hyperparameters, entropy weight $\alpha=0.01$, ensemble size $M=64$, LCB coefficient $\beta=4$ and diffusion steps  $T=5$ for all tasks in different domains. As for others related to our algorithm, we consider the policy learning rate, Q-learning weight $\eta$,  and whether to use max Q backup.  For implementation details, we consider the gradient clip norm, diffusion loss type, and whether to clip action at every diffusion step. We keep the hyperparameter same for tasks in the same domain except for the AntMaze domain. We use max Q backup \cite{kumar2020conservative} for complete tasks.
The hyperparameter settings are shown in Table~\ref{tab:hyperparam}.
% \small
\begin{table*}[ht]
\centering
\caption{Hyperparameter settings of all selected tasks. `*' means all the AntMaze tasks use max Q-backup trick~\cite{kumar2020conservative} except the `antmaze-umaze-v0' task as the same as that in other papers. The `likelihood' loss is proposed in IR-SDE \cite{luo2023image}  which forces the model to learn optimal reverse paths from $a^t$ to $a^{t-1}$.}
\label{tab:hyperparam}
\resizebox{1.\linewidth}{!}{
\begin{tabular}{l|ccccccc}
\toprule
Tasks domain    &    learning rate  &   $\eta$  &     max Q-backup & gradient norm  &  loss type &   action clip & $T$\\
\midrule
Gym &    3e-4 &   1.0  &     False & 4.0 &    Likelihood&   False & 5\\
% hopper-medium-v2&    3e-4 &   1.0  &     False & 4.0 &    Likelihood&   False\\
% walker2d-medium-v2    &    3e-4 &   1.0  &     False & 4.0 &    Likelihood&   False\\
% halfcheetah-medium-replay-v2&    3e-4 &   1.0  &     False & 4.0 &    Likelihood&   False\\
% hopper-medium-replay-v2     &    3e-4 &   1.0  &     False & 4.0 &    Likelihood&   False\\
% walker2d-medium-replay-v2   &    3e-4 &   1.0  &     False & 4.0 &    Likelihood&   False\\
% halfcheetah-medium-expert-v2&    3e-4 &   1.0  &     False & 4.0 &    Likelihood&   False\\
% hopper-medium-expert-v2     &    3e-4 &   1.0  &     False & 4.0 &    Likelihood&   False\\
% walker2d-medium-expert-v2   &    3e-4 &   1.0  &     False & 4.0 &    Likelihood&   False\\
% \midrule
AntMaze&    3e-4 &   2.0  &     True* & 4.0 &  Noise    &   True & 5\\
% antmaze-umaze-v0&    3e-4 &   2.0  &     False & 4.0 &  Noise    &   True\\
% antmaze-umaze-diverse-v0    &    3e-4 &   2.0  &     True  & 4.0 &  Noise    &   True\\
% antmaze-medium-play-v0&    3e-4 &   2.0  &     True  & 4.0 &  Noise    &   True\\
% antmaze-medium-diverse-v0   &    3e-4 &   2.0  &     True  & 4.0 &  Noise    &   True\\
% antmaze-large-play-v0 &    3e-4 &   2.0  &     True  & 4.0 &  Noise    &   True\\
% antmaze-large-diverse-v0    &    3e-4 &   2.0  &     True  & 4.0 &  Noise    &   True\\
% \midrule
Adroit    &    3e-5 &   0.1  &     False &8.0 &  Noise    &   True& 10\\
% pen-human-v1    &    3e-5 &   0.1  &     False &8.0 &  Noise    &   True\\
% pen-cloned-v1   &    3e-5 &   0.1  &     False &8.0 &  Noise    &   True\\
% \midrule
Kitchen   &    3e-4 &   0.005&     False &10.0&     Likelihood&   False& 10\\
% kitchen-complete-v0   &    3e-4 &   0.005&     False &10.0&     Likelihood&   False\\
% kitchen-partial-v0    &    3e-4 &   0.005&     False &10.0&    Likelihood&   False\\
% kitchen-mixed-v0&    3e-4 &   0.005&     False &10.0&    Likelihood&   False\\
\bottomrule
\end{tabular}
}
\end{table*}
% \normalsize

\subsection{Automating Entropy Adjustment}
\label{app-sec:auto-alpha}
\begin{figure}
    \centering
    \begin{minipage}{\textwidth}
    \centering
    \includegraphics[width=0.55\linewidth]{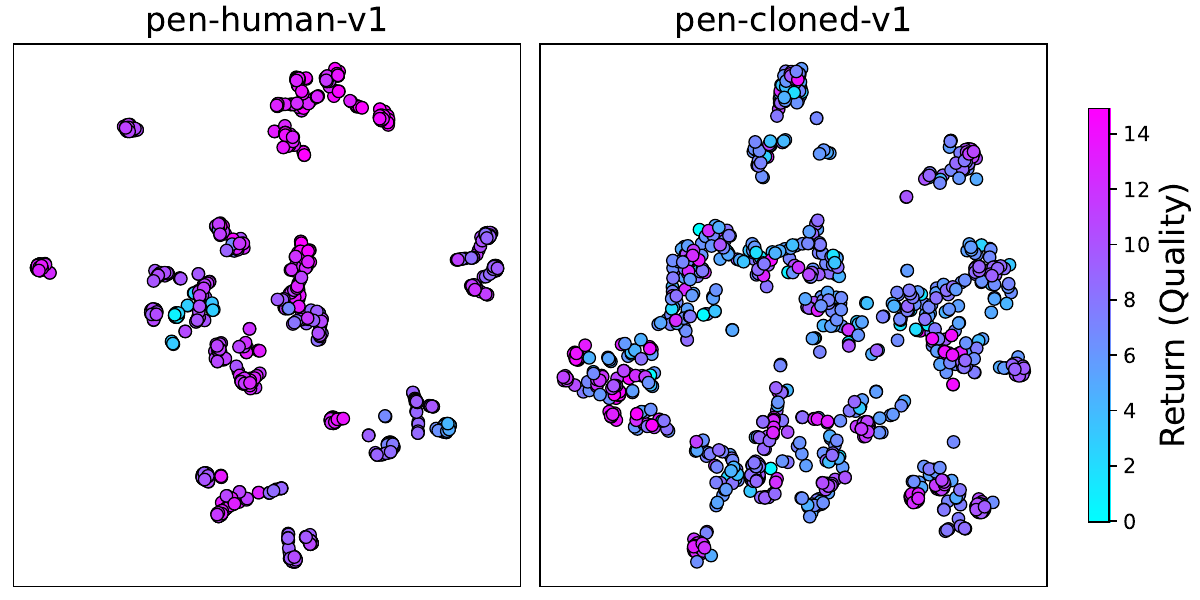}
    % \caption{(a) Adroit tasks}
    \label{fig:tsne-pen}
    \end{minipage}
    
    \begin{minipage}{\textwidth}
    \centering
    \includegraphics[width=0.7\linewidth]{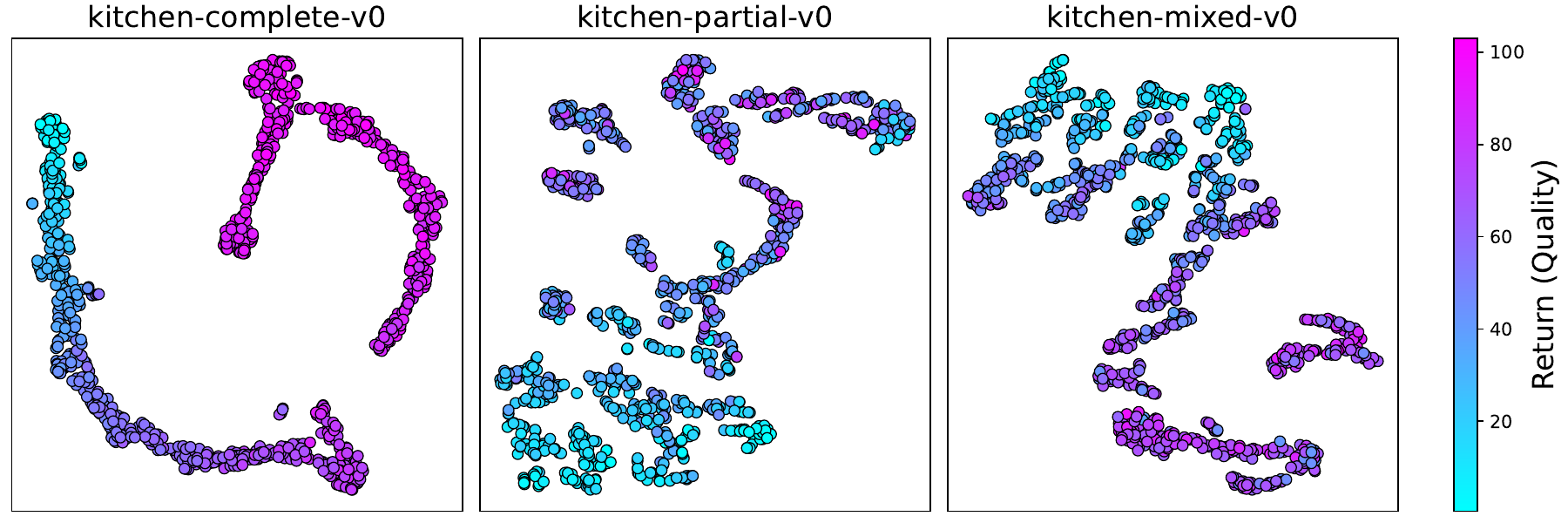}
    % \caption{(b) Kitchen tasks}
    \label{fig:tsne-kitchen}
    \end{minipage}

    \begin{minipage}{\textwidth}
    \centering
    \includegraphics[width=\linewidth]{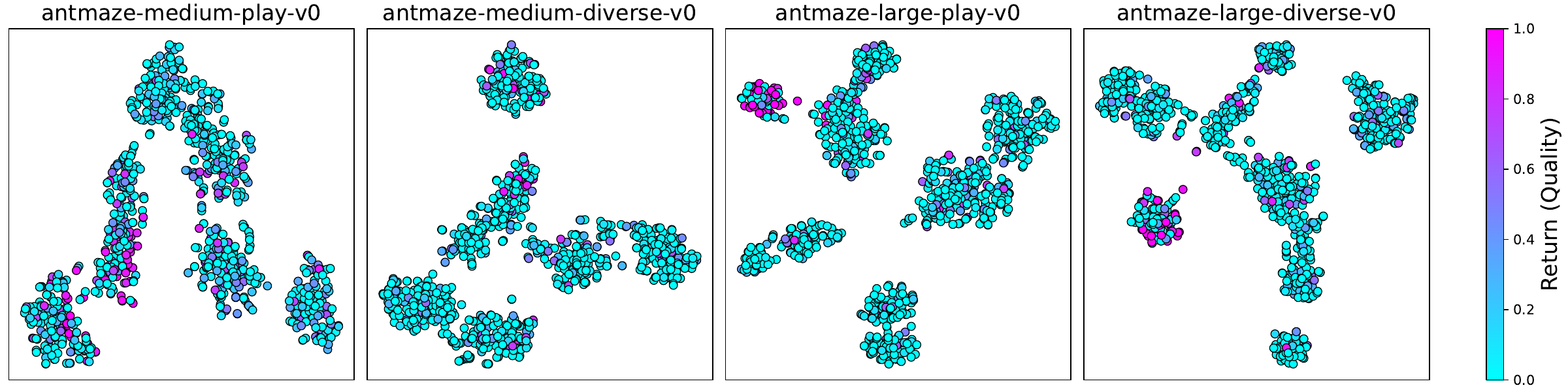}
    % \caption{(c) Antmaze tasks}
    \label{fig:tsne-antmaze}
    \end{minipage}
    \caption{A t-SNE visualization of randomly selected 1000 states from Antmaze, Adroit and Kitchen domain. The color coding represents the return of the trajectory associated with each state.}
    \label{fig:tsne}
\end{figure}

It is possible to consider automating entropy adjustment similar to \citep{haarnoja2018soft} but $\alpha$ depends on the state since
\blue{
the offline dataset is pre-collected and may be imbalanced across different states shown in Figure~\ref{fig:tsne}. 
}
One way to compute the gradients for state-depend $\alpha$ is 
\begin{equation}
    J(\alpha) = \mathbb{E}_{s_i \sim \mathcal{D}, a_i\sim \pi_\phi}[-\alpha(s_i)\log \pi_\phi (a_i\vert s_i) - \alpha(s_i) \mathcal{\bar{H}}],
\end{equation}
where \blue{$\alpha(s_i)$ is implemented as a neural network with a single hidden layer consisting of 32 units} and $\mathcal{\bar{H}}$ is a desired minimum expected entropy which represents the desired level of exploration which can be set as a function of action space dimension without the need for extensive hyperparameter tuning across different tasks. 

\subsection{More Analysis for Q-Ensembles}
\begin{figure}[ht]
    \centering
    \includegraphics[width=\textwidth]{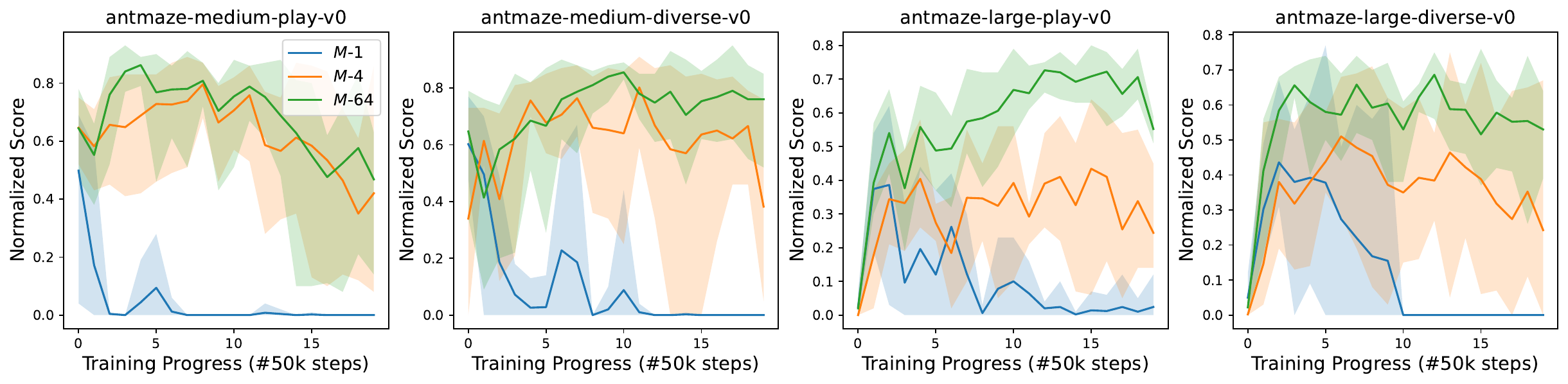}
    \caption{Abaltion study of Q-ensemble size $M$ on selected AntMaze tasks. We consider $M\in \{1,4,64\}$and we found size $64$ is the best overall tasks.}
    \label{app-fig:abla-M}
\end{figure}

Here, we provide more detailed experiments for analyzing the effect of ensemble sizes $M$ as we discussed in Section~\ref{sec:analysis}. More specifically, the results of different ensemble sizes are reported in Table~\ref{app-table:q-ensemble} and Figure~\ref{app-fig:abla-M}, in which we also provide the variance that further shows the robustness of our method. 

\tiny
\begin{table*}[ht]
\centering
\caption{Ablation study of ensemble size $M$ on selected AntMaze tasks. }
\label{app-table:q-ensemble}
\begin{center}
% \begin{small}
% \begin{sc}
\resizebox{1.\linewidth}{!}{
\begin{tabular}{l|c|c|c|c|c}
\toprule
   Ensemble Size    & $M=1$      & $M=2$   & $M=4$       & $M=16$  & $M=64$ \\
\midrule
antmaze-medium-play-v0   & $50.2 \pm 26.4$   & $84.0 \pm 5.0$  & $87.2 \pm 1.1$ & $83.6\pm 7.7$    &$\textbf{91.4} \pm 1.5$\\
antmaze-medium-diverse-v0   & $67.2 \pm 7.6$ & $71.8 \pm 14.1$   & $87.2 \pm 3.8$ & $88.0\pm 2.2$     &$\textbf{91.6} \pm 2.3$\\
antmaze-large-play-v0       & $48.2 \pm 10.8$ & $54.2 \pm 10.0$  & $52.4 \pm 13.0$& $71.8\pm 5.8$ &$\textbf{81.2} \pm 3.0$\\ 
antmaze-large-diverse-v0     & $58.8 \pm 11.4$  & $43.2 \pm 16.1$  & $69.0 \pm 8.3$ & $76.4 \pm 8.47$ &
$\textbf{76.4} \pm 2.1$\\
\midrule
\textbf{Average}         & 56.1     & 63.3  & 74.0  & 80    &\textbf{85.2}\\
\bottomrule
\end{tabular}
% \end{sc}
% \end{small}
}
\end{center}
\end{table*}

\normalsize

\subsection{Ablation study on LCB coefficients $\beta$}
To explore the impact of different LCB coefficients $\beta$. We add an experiment of our method with $\beta$ values of 1, 2, and 4 on AntMaze-medium environments. Figure~\ref{fig:lcb-beta} demonstrates that adjusting the LCB coefficient improves performance, particularly for higher values, which helps in managing the exploration-exploitation trade-off effectively. In addition, the numerical results are provided in Table~\ref{table:lcb}.

\begin{table}[ht]
    
% \begin{wraptable}{r}{0.5\textwidth}
\vspace{-1.3em}
\centering
% \scriptsize
\caption{Ablation study of LCB coefficients $\beta$.}
\label{table:lcb}
% std
% \begin{center}
% \begin{small}
% \begin{sc}
% \vspace{-2mm}
% \resizebox{1.\linewidth}{!}{
\begin{tabular}{lccc}
\toprule
   LCB Coefficient   $\beta$    & $1$   & $2$    &  $4$ \\
\midrule
Antmaze-medium-play-v0   & $82.4 \pm 4.9$ & $88.6 \pm 1.5$ & \textbf{91.6}$\pm 2.3$     \\
Antmaze-medium-diverse-v0      & $74.6\pm 3.7$  & $84.0\pm 7.8$  & \textbf{91.4} $\pm 1.5$    \\
% Antmaze-large-play-v0       &   & $81.2 \pm 3.0$  & \\ 
% Antmaze-large-diverse-v0      & $81.0 \pm 7.0$   & $76.4 \pm 2.1$ & \\
% Kitchen-complete-v0 \\
% Kitchen-Mixed-v0 &  \\
\midrule
\textbf{Average}            &  78.5 &   86.3  & \textbf{91.5}\\
\bottomrule
\end{tabular}
% }
% \end{sc}
% \end{small}
% \end{center}
% \vspace{-2em}
% \end{wraptable}

\end{table}

\subsection{Ablation study on Diffusion step $T$}
\label{app-sec:ablation-T}
We evaluated the impact of varying the number of diffusion steps on a range of tasks, including AntMaze, Gym, and Kitchen in Table~\ref{table:T}. Our findings indicate that while increasing the number of steps generally improves performance, five steps provide the best balance across different tasks and between performance and computational time in Gym and Antmaze tasks. For more complex tasks as in Kitchen and Pen, we choose $T=10$.
\begin{table}[ht]
    
% \begin{wraptable}{r}{\textwidth}
% \vspace{-1.3em}
% \vspace{-1em}
\centering
% \scriptsize
\caption{Ablation study of diffusion step $T$.}
\label{table:T}
% \begin{center}
% \begin{small}
% \begin{sc}
% \vspace{-2mm}
% \resizebox{1.\linewidth}{!}{
\begin{tabular}{lccc}
\toprule
   Diffusion Step   $T$    & $3$   & $5$    &  $10$ \\
\midrule

Halfcheetah-medium-replay-v2& 43.4 & \textbf{57.0} & 49.5 \\
Hopper-medium-replay-v2& 39.4 & \textbf{102.7} & 101.7 \\
Walker2d-medium-replay-v2& 51.2 & 94.2 & \textbf{98.1} \\

\midrule
Antmaze-medium-play-v0   & \textbf{96.6}   & $91.6$     &$90.2$\\
% Antmaze-medium-play-v0   & \textbf{96.6} $ \pm 2.7$  & $91.6\pm 2.3$     &$90.2\pm 1.8$\\
Antmaze-medium-diverse-v0      & \textbf{95.8} & $91.4$     &$83.8 $\\
% Antmaze-medium-diverse-v0      & \textbf{95.8} $ \pm 1.5$ & $91.4\pm 1.5$     &$83.8 \pm 5.4$\\
Antmaze-large-play-v0       &  67.6  & \textbf{81.2}  & $63.2$ \\ 
% Antmaze-large-play-v0       &  67.6 $\pm 6.6$ & $81.2 \pm 3.0$  & \\ 
Antmaze-large-diverse-v0      & \textbf{81.0}   & 76.4  & 70.0 \\
% Antmaze-large-diverse-v0      & $81.0 \pm 7.0$   & $76.4 \pm 2.1$ & \\
\midrule
pen-human-v1 & 65.4 & 67.2 & \textbf{70.0} \\
pen-cloned-v1 & 67.3  & 66.3 & \textbf{68.4}\\
\midrule
Kitchen-complete-v0 & 7.5 & 82.3 & \textbf{92.7}  \\
% Kitchen-complete-v0 & 7.5 $\pm5.4$ & 82.3 & \textbf{92.7} $\pm$ 1.9 \\
Kitchen-partial-v0 &  10.9  & 60.3  & \textbf{66.3}\\
% Kitchen-partial-v0 &  10.9 $\pm 7.6$  & 60.3 $\pm 6.2$ \\
Kitchen-mixed-v0 &  4.8  & 60.2  & \textbf{68.0}\\
% Kitchen-mixed-v0 &  4.8 $\pm 6.6$  & 60.2 \\
% 4.8 $\pm 6.6$
\midrule
\textbf{Average}            & 52.6  &  \textbf{77.6}   &  76.8\\
\bottomrule
\end{tabular}
% }
% \end{sc}
% \end{small}
% \end{center}
\vspace{-1em}
% \end{wraptable}

\end{table}

\subsection{Ablation study on Max Q-back}
We conducted experiments with and without max Q-backup on AntMaze tasks in Table~\ref{table:max-q}. The inclusion of max Q-backup significantly enhances performance, particularly in more complex environments (e.g., Antmaze-large).
\begin{table}[ht]
    
% \begin{wraptable}{}{\textwidth}
% \begin{wraptable}{l}{\textwidth}
\vspace{-0.2em}
\centering
% \scriptsize
\caption{Ablation study of "Max Q trick".}
\label{table:max-q}
% \begin{center}
% \begin{small}
% \begin{sc}
% \vspace{-2mm}
% \resizebox{1.\linewidth}{!}{
\begin{tabular}{lcc}
\toprule
   Max Q-backup      & True   & False \\
\midrule
Antmaze-medium-play-v0   & \textbf{91.6}$\pm 2.3$ & $89.2 \pm 2.9$ \\
Antmaze-medium-diverse-v0      & \textbf{91.4}$\pm 1.5$  & $87.6\pm 1.8$\\
Antmaze-large-play-v0       & \textbf{81.2} $\pm 3.0$ & 22.3 $\pm$ 7.1\\ 
Antmaze-large-diverse-v0      & \textbf{76.4} $\pm 2.1$  & 26.5 $\pm$ 6.1\\
% \midrule
% \textbf{Average}            &     & \\
\bottomrule
\end{tabular}
% }
% \end{sc}
% \end{small}
% \end{center}
% \vspace{-1em}
% \end{table}
% \end{wraptable}

\end{table}

\subsection{Offline vs Online Model Selection}
We use the online experience to evaluation our model during training. 
Table~\ref{tab:online_vs_offline} presents a comparison of our method with Diffusion-QL, including both online and offline results. Additionally, we include our method's performance based on offline selection using the BC Loss criterion, selecting the step where the difference between consecutive steps was less than 4e-3.
\begin{table}[]

    \centering
    \caption{Performance comparison with online model selection and offline model selection. }
    \resizebox{1.\linewidth}{!}{
    \begin{tabular}{l|cccc}
        \toprule
        AntMaze Tasks &  Diffusion-QL (Offline) &  DIffusion-QL (Online) & Ours (Offline)& Ours (Online)\\
        \midrule
        antmaze-umaze-v0 & 93.4 & 96.0 &  99.0  & \textbf{100.0} \\
        antmaze-umaze-diverse-v0 & 66.2& \textbf{84.0}  &  67.5  &79.8\\
        antmaze-medium-play-v0 & 77.6 & 79.8 &  84.0  &\textbf{91.4}\\
        antmaze-medium-diverse-v0 & 78.6 & 82.0  &  85.4  &\textbf{91.6}\\
        antmaze-large-play-v0 & 46.6 & 49.0 &  72.6  &\textbf{81.2}\\
        antmaze-large-diverse-v0 & 56.6 & 61.7 &  65.9  &\textbf{76.4}\\
        \midrule
        Average & 69.6 & 75.4 &  79.2  & \textbf{86.7} \\
        \bottomrule
    \end{tabular}
    }
    \label{tab:online_vs_offline}
\end{table}

\newpage
%%%%%%%%%%%%%%%%%%%%%%%%%%%%%%%%%%%%%%%%%%%%%%%%%%%%%%%%%%%%

%%%%%%%%%%%%%%%%%%%%%%%%%%%%%%%%%%%%%%%%%%%%%%%%%%%%%%%%%%%%

% \newpage
% \newpage
% \input{neurips_check_list}

\end{document}